# HLS-GPT: A Generative Pretrained Transformer (GPT) for Continental-Scale NASA Harmonized Landsat and Sentinel-2 (HLS) Reflectance Reconstruction Across All Bands on Arbitrary Dates

Junjie Li [a], Hankui K. Zhang [a, *], David P. Roy [b]

a, Geospatial Sciences Center of Excellence, Department of Geography and Geospatial Sciences, South Dakota State University, Brookings, SD 57007, USA b, Department of Geography, Environment, and Spatial Sciences, & Center for Global Change and Earth Observations, Michigan State University, MI 48824, USA

**Abstract**:

Promising deep learning–based Landsat and Sentinel-2 reflectance time series reconstruction methods have been developed recently but are restricted to select sensor bands, lack demonstrated geographic scalability, and/or are limited by patch-based designs with short temporal coverage that under-utilize surface seasonal dynamics. We present a new large-scale generative pre-trained Transformer (GPT) model that reconstructs NASA Harmonized Landsat Sentinel (HLS) 30 m surface reflectance for all bands, any date, and pixel location. The developed HLS-GPT model uses a hierarchical Transformer structure to accommodate the different number of spectral bands sensed by Landsat and Sentinel-2 and is applied to any single pixel location 12-month time period. To capture geographic and temporal surface variability, the model was trained using nine years of single-pixel HLS reflectance time series extracted at >0.25 million conterminous United States (CONUS) training pixel locations. A novel random cropping and masking training approach is used where 12-month time periods are extracted with random start days that vary among training epochs for each training pixel location; 50% of the observations are randomly masked and the model is trained to reconstruct their reflectance from the unmasked observation reflectance values. CONUS results considering >62,000 independent test pixel locations is shown to provide robust reconstruction under diverse conditions including complex crop phenology and for sparse and irregular time series. Leave-one-observation-out test data evaluation provides all HLS spectral band reconstruction RMSE <0.026, with relative RMSE <35% for the visible bands (most sensitive to residual atmospheric contamination) and <13% for the other bands. In addition, red-edge band RMSE values are comparable to red and NIR RMSE values despite Landsat acquitting no red-edge band. Sensitivity analyses using random masking of a fixed proportion (10%, 20%, …, 90%) of the test data time series provided only slightly worse all band RMSE (<0.028) and relative RMSE (<36% for the visible bands, <14% for the other bands) when from 10% to 50% of the test data observations were masked. Image reconstruction for nine 109 × 109 km CONUS HLS tiles not included in the training demonstrates the model's superiority over two conventional methods and the recent NASA-IBM Prithvi model.

## 1. Introduction

Global coverage medium resolution satellite data are currently acquired by the ESA Sentinel-2 Multispectral Imager (MSI) (Drusch et al., 2012) and by the NASA/USGS Landsat-8/9 Operational Land Imager (OLI) (Roy et al., 2014a; Masek et al., 2020) that are in polar sun-synchronous morning overpass time orbits with individual sensor revisits of 10 and 16 days respectively. Higher temporal resolution observations are needed to enable new and improved land and water, management, decision-support, and monitoring applications (Radeloff et al., 2024) and have been advocated for by Landsat science teams and earth observation community surveys (Wu et al., 2019; Roy et al., 2026). In recognition of this need, the NASA Harmonized Landsat and Sentinel-2 (HLS) product provides global coverage 30 m reflectance processed to minimize among sensor differences with higher temporal resolution than either Landsat or Sentinel-2 alone (Ju et al., 2025). For example, in 2022 there was an average of 69 good-quality cloud-free HLS observations per 30 m global land pixel, equivalent to better than weekly temporal resolution (Zhou et al., 2025). Most applications, including, for example, recent HLS-based land surface phenology detection (Tran et al., 2025), within-season crop mapping (Zhang et al., 2025a), and disturbance and land change monitoring (Shang et al., 2022; Pickens et al., 2025), would be improved by the availability of daily or near-daily medium resolution reflectance observations provided with no gaps. However, polar-orbiting sun-synchronous reflectance observations are not daily and are typically irregular in time. For the HLS data this is because the Landsat and Sentinel-2 orbit swaths overlap with geographically complex patterns (Li and Roy, 2017), clouds preclude surface observation and globally some regions and seasons are considerably more cloudy than others at morning overpass times (Kovalskyy and Roy, 2013), and because the HLS data have chronologically increasing observation density due to the inclusion of new observations as new satellites were launched, starting with Landsat-8 (2013), Sentinel-2A (2015), Sentinel-2B (2017), Landsat-9 (2021), and Sentinel-2C (2024) (Ju et al., 2025).

A significant body of research has been undertaken over the past several decades to develop algorithms that provide gap-free reflectance Landsat and more recently Sentinel-2 time series. We refer to them generically for convenience as reconstruction algorithms. Reconstruction algorithms can be broadly categorized into ones that use or do not use additional auxiliary satellite data. Early reconstruction algorithms were developed for Landsat using higher temporal but lower spatial resolution auxiliary MODIS data, including the widely used STARFM (Gao et al., 2006) and algorithm variants (e.g., Zhu et al., 2016). This approach was also implemented for Sentinel-2 time series reconstruction using higher temporal but lower spatial resolution auxiliary Sentinel-3 data (Wang and Atkinson, 2018; Guo et al., 2024). Optical wavelength–SAR data fusion methods, including DSen2-CR (Meraner et al., 2020), GLF-CR (Xu et al., 2022a), ST-MLR (Mao et al., 2023), PIGAN (Dai et al., 2025), and RESTORE-DiT (Shu et al., 2025) were developed to exploit the cloud-penetrating capability of SAR data for the reconstruction of cloud-contaminated optical wavelength data. However, spatial and spectral resolution differences between the auxiliary and the optical wavelength data may limit the practical applicability of these reconstruction methods (Roy et al., 2008; Wu et al., 2024; Michel & Inglada, 2026).

Reconstruction approaches that do not rely on auxiliary satellite data sources directly model the temporal dynamics and/or the spatial relationships of reflectance data from the sensor data. Temporal dynamics–based approaches were first developed for coarse-resolution satellite data in the early 2000s and have more recently been adapted for Landsat and Sentinel-2. These include Harmonic Analysis of Time Series (HANTS), a harmonic function–based model (Roerink et al., 2000) and its variants (Zhou et al., 2015; Zhu et al., 2015; Roy and Yan, 2020), Savitzky–Golay filter (Chen et al., 2004; Chen et al., 2021), asymmetric Gaussian models (Jonsson & Eklundh, 2002; Shao et al., 2016; Hurley et al., 2014), and double logistic functions (Zhang et al., 2003; Liu et al., 2017; Wu et al., 2021). Other approaches combine temporal dynamics with spatial similarity among neighboring pixels for reconstruction, including spatiotemporal extrapolation/interpolation (Weiss et al., 2014; Yan and Roy, 2020; Wu et al., 2024), statistical and machine learning methods (Cao et al., 2020; Song et al., 2025), and optimization-based methods such as dictionary learning (Li et al., 2014) and tensor completion (Chu et al., 2021). Temporal dynamics models are intuitive and interpretable for describing seasonal variations in land surface dynamics and have been applied for large area applications. However, they have strong assumptions of signal stationarity, periodicity, and symmetry, are less able to accommodate abrupt changes (e.g., due to drought, flood, fire, deforestation), or complicated surface changes (e.g., alfalfa with multiple harvests in a year), and are less reliable when the observations are temporally sparse or irregular, or if several years of data are used and there is inter-annual variability in the surface cover and/or condition (Brooks et al., 2012; Roy and Yan, 2020).

In recent years, deep learning-based approaches have been developed with impressive reconstruction performance. A summary of recent deep learning-based single-source reconstruction studies is provided in Appendix A including their input–output design (whether patch-based or single-pixel–based), training data size, training sequence length, reconstructed bands, and publication status. Most existing approaches adopt a patch-based framework (Chen et al., 2020; Stucker et al., 2023; Szwarcman et al., 2025; Wang et al., 2025a), with typical patch sizes ranging from 40 × 40 to 256 × 256 pixels, largely reflecting the influence of deep learning models originally developed for computer vision tasks. Acquiring cloud-free patches through time at the same location remains challenging, and both sequence length and patch size substantially increase computational cost (Song et al., 2025). As a result, most models are trained with limited temporal coverage (often fewer than 10 time steps), restricting their ability to capture complex intra-annual dynamics and full seasonal patterns that are typically exploited by conventional approaches. Recent approaches that model seasonal patterns using single-pixel time series (Liu et al., 2024; Zhang et al., 2025b) have been developed but are trained and applied to fixed 12-month (January to December) calendar year periods. This design limits their general applicability and prevents temporally rolling implementation, which is particularly important for near-real-time applications. Most approaches reconstruct only a subset of the 7 and 11 spectral bands acquired by Landsat and Sentinel-2 respectively, and commonly reconstruct only the red, green and blue visible bands, reflecting their computer vision algorithm origins. In addition, most approaches have been conducted over relatively small geographic regions, limiting model generalization and scalability for national-scale reconstruction. Further, for many studies the trained model is not provided and users must train the model from scratch, which can be computationally demanding and labor-intensive. A notable exception is the patch-based NASA–IBM model, which was trained using global-scale data, although it still suffers from the limitations noted above, including limited temporal dynamics and restricted spectral band reconstruction.

To address the aforementioned limitations, we propose HLS-GPT, a generative pre-trained Transformer for accurate reconstruction of HLS reflectance time series across all spectral bands and developed for application to one-year data sequences beginning on any arbitrary date. The goal of HLS-GPT is to provide a ready-to-use pre-trained model that enables users to reconstruct all seven Landsat and eleven Sentinel-2 reflectance bands for any pixel at any date across the conterminous United States (CONUS), without requiring additional auxiliary data or task-specific retraining. Specifically, a hierarchical Transformer architecture is used with two independent Transformers to process the Landsat and Sentinel-2 time series and accommodate their different numbers of spectral bands. A third Transformer is used to integrate concatenated sensor features similar to the approach we developed for HLS-based within-season crop type classification (Zhang et al., 2025a). The HLS-GPT model was trained using spatial (96 109 × 109 km HLS tiles, out of 976 CONUS tiles) and temporal (9 years) HLS CONUS data. At each training pixel location, 12-month time periods were extracted with random start days from each of the first eight years. The start dates were varied among training epochs and among pixel locations to increase training diversity. For each 12-month period, 50% of the valid observations were randomly masked and the model was trained to reconstruct their reflectances from the unmasked values. To evaluate the reconstruction performance of HLS-GPT, two different validation strategies were conducted and tile-level assessments on a held-out test dataset as well as nine independent HLS tiles are presented. The results were further compared with two conventional non-deep learning time series fitting methods (Roerink et al., 2000; Chen et al., 2004) and the recent NASA–IBM Prithvi version 2.0 model (Szwarcman et al., 2025). The paper finishes with a discussion of the proposed HLS-GPT model and its potential applications.

## 2. Data

### 2.1 The NASA Harmonized Landsat and Sentinel-2 (HLS) product

The HLS product provides global atmospherically corrected nadir BRDF adjusted (NBAR) 30 m reflectance derived from the Landsat-8/9 and Sentinel-2A/B/C data processed to minimize sensor differences (Ju et al., 2025). At CONUS and other mid-latitudes, a location may be sensed on the same day by Landsat and Sentinel-2, but no location is sensed on the same day by Landsat-8 and Landsat-9, and no location is sensed on the same day by Sentinel-2A and Sentinel-2B. The HLS processing includes atmospheric correction using the same radiative transfer model and atmospheric characterization for each sensor (Vermote et al., 2018), image-based cloud masking (Qiu et al., 2019), adjustment of reflectance to NBAR using the *c*-factor method (Roy et al., 2016), and radiometric adjustment of the Sentinel-2 spectral bands to the Landsat OLI bands. The HLS processed Landsat-8 (since 2013) and Landsat-9 (since 2021) seven reflective wavelength bands (four visible, one NIR and two SWIR), were used. The HLS processed Sentinel-2A (since 2015) and Sentinel-2B (since 2017) bands that have similar seven reflective wavelength bands as Landsat and in addition a broad band NIR and three red-edge bands were used. The seven band HLS Landsat and eleven band HLS Sentinel-2 30 m atmospherically corrected NBAR data were used. For convenience we refer to the surface NBAR data as reflectance hereafter.

The HLS data includes per-pixel quality and cloud flags (Ju et al., 2025). Only good quality HLS pixel observations (flagged as cloud and shadow-free, unsaturated, snow-free, and not adjacent to cloud or cloud shadow) were used. In addition, negative HLS reflectance values, caused by atmospheric "overcorrection" associated primarily with incorrect atmospheric characterization (Roy et al., 2014b), were discarded. The HLS Landsat and Sentinel-2 reflectance data are not perfectly consistent due to sensor calibration errors, residual among sensor spectral band pass differences, error from the reflectance adjustment to NBAR, and, in particular due to atmospheric correction errors (Ju et al., 2025). Global analysis of same-day Landsat and Sentinel-2 HLS observation pairs indicate that the corresponding sensor red, NIR, and the two SWIR reflectance bands had differences equivalent to 4% of the mean Landsat surface reflectance and up to 10.5% for the green, blue and coastal blue bands (Ju et al., 2025). This is important for contextualizing the HLS-GPT reconstruction results described later below.

The HLS data are provided in fixed spatially overlapping 109 × 109 km tiles in the Universal Transverse Mercator (UTM) map projection in the same Military Grid Reference System (MGRS) tiling system used to store ESA Sentinel-2 data (Bauer-Marschallinger and Falkner, 2023).

Consequently, the HLS processed Sentinel-2 data are down-sampled from their native resolutions to 30 m, and the HLS processed 30 m Landsat data are reprojected to the MGRS tiles (Ju et al., 2025). Each HLS tile is composed of 3660 × 3660 30 m pixels and is referenced with a unique Tile ID (HLS, 2025). In this study the HLS data for nine years (January 1st 2015 to December 31st 2023) composed of Landsat-8, Landsat-9, Sentinel-2A and Sentinel-2B processed data for 96 training/test (Section 2.2) and 9 independent evaluation (Section 2.3) CONUS HLS tiles were used.

## 2.2 HLS-GPT training and test time series data

The HLS-GPT model was trained and tested using single pixel good quality reflectance time series extracted from 96 HLS tiles distributed systematically across the CONUS that contain diverse land cover (Fig. 1, black boundaries). For each tile, single-pixel 9-year HLS times series were extracted every 60 pixels in the tile $x$ and $y$ axes providing a total of 312,950 pixel locations that were randomly divided by pixel location 80% as training (250,360 pixel locations) and 20% as test (62,590 pixel locations). A total of >130 million good quality observations (>104 million training and >26 million test) were extracted over the 9 years and 35% were sensed by a Landsat sensor and 65% by a Sentinel-2 sensor.

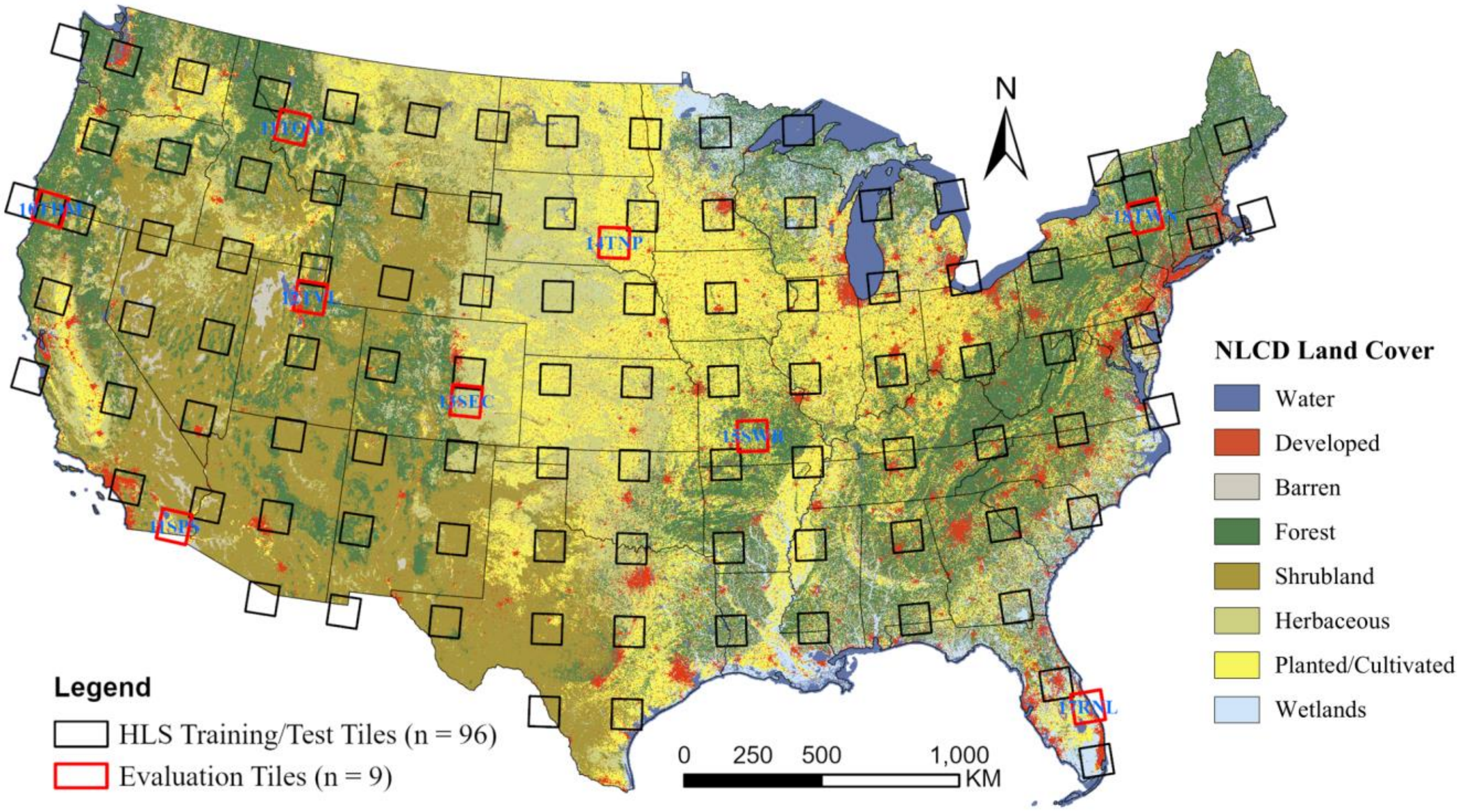

**Fig. 1.** Locations of the 96 HLS 109 × 109 km tiles (black boundaries) used to extract HLS-GPT training and test data, and the locations of 9 HLS tiles used for the independent reconstruction evaluation (red boundaries) show superimposed on the year 2023 eight class Level-1 National Land Cover Database (NLCD) 30 m land cover product (Sohl et al., 2025).

Fig. 2 shows boxplots of the annual number of days with good-quality observations for each of the nine years, illustrated separately for the Landsat (blue), Sentinel-2 (orange) and both sensors (black), derived considering all the CONUS training pixel locations. The Landsat median count (50$^{th}$ percentile, solid horizontal line) values varied from 11 (2019) to 24 (2022) days whereas the median Sentinel-2 counts were lower in 2015 and 2016, as only Sentinel-2A data were in the HLS dataset in those years, but increased in the following years with the introduction of Sentinel-2B data with median counts varying from 15 (2017) to 38 (2021) days. When considering both sensors together the median counts vary from 12 (2015) to 60 (2022) days. Some CONUS training pixel locations had as many as 128 (Sentinel-2), 81 (Landsat), and 176 (both sensors) days with good-quality HLS observations in a year, and occurred predominantly in the south-west where cloud cover at the time of overpass is low and adjacent satellite orbits overlap (Egorov et al., 2019; and Fig. 1 in Zhang et al., 2023).Some locations had days with no good-quality Sentinel-2 or Landsat observations throughout the year, primarily due to persistent cloud or cloud shadow cover identified by the HLS quality and cloud flags.  However, considering all nine years, in any year at least 97.8% of the CONUS training pixels locations had ≥8 good-quality observations. The test data had similar annual patterns. The Fig. 2 statistics capture the variability of the HLS observation data through time and emphasize the need for caution in interpreting science results derived from annual HLS time series without considering the amount of time series data available.

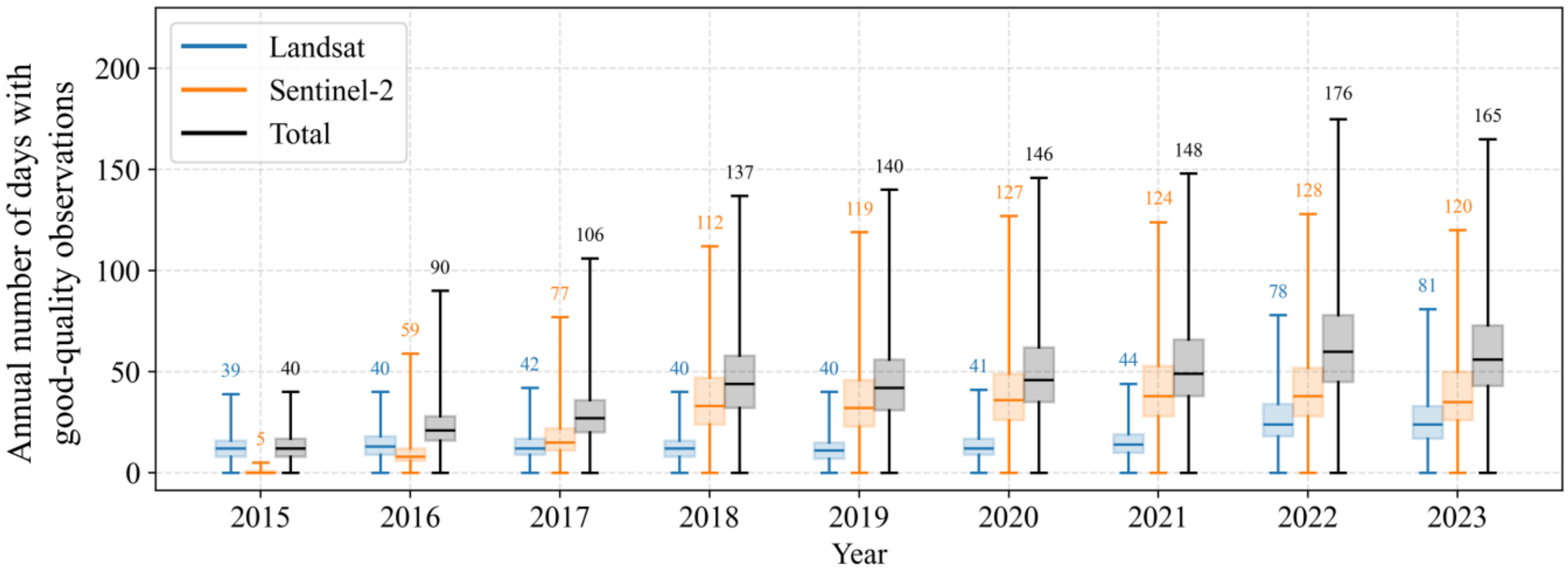

**Fig. 2.** Summary statistics (minimum, 25$^{th}$ percentile, 50$^{th}$ percentile i.e. median, 75$^{th}$ percentile, and maximum) of the annual number of days with good-quality Landsat, Sentinel-2 and both sensor, HLS observations for the 250,360 30 m CONUS training pixel locations. Note that a minimum of zero indicates that some pixels had days with no good-quality observations in a given year, mainly due to persistent cloud/snow contamination indicated by the HLS quality and cloud flags.

## 2.3 HLS independent evaluation tiles

To demonstrate the practical applicability of the HLS-GPT model, reconstruction results for nine HLS tiles (Fig. 1, red boundaries) that do not overlap with the 96 training/test tiles, and so provide an independent evaluation, were considered. The independent evaluation was undertaken using 2023 HLS data and salient tile information for that year is summarized in Table 1. Year 2023 was selected as it has a representative amount of data acquired by two Landsat and two Sentinel-2 sensors (Fig. 2). Of the 9 tiles, 5 were used in our previous Liu et al. (2024) study where they are described in detail, and include tiles that have a high (southern California, TileID 11SPS) and low (upstate New Yori, TileID 18TWN) number of cloud-free observations, a predominantly agricultural tile in South Dakota (TileID 14TNP), a coastal tile in Florida (TileID 17RNL), and a mountainous tile in Utah (TileID 12TVL). Four additional tiles were selected in Montana (TileID 11TQM), Colorado (TileID 13SEC), Missouri (TileID 15SWB), and northern California (TileID 10TDM) that includes the Smith River Complex wildfire (USFS, 2023).

**Table 1**. Summary of the 9 independent evaluation HLS tiles: Tile ID, U.S. State, top three land cover classes defined by the 2023 NLCD product (percentage of tile pixels in each class shown in parentheses), mean tile cloud cover for 2023 as defined by the HLS data, mean number of year 2023 good-quality observations averaged over all the tile pixels, and the year 2023 reconstruction target date and sensor that acquired data on that date (described in Sections 3.5 and 4.3).

| Tile ID | State(s) | Dominant year 2023 NLCD land cover type (top 3) | Mean tile cloud cover for 2023 as defined by the HLS data | Mean number of good quality HLS tile observations for 2023 | Reconstruction target date day of year in 2023 (sensor acquisition on day) |
|---|---|---|---|---|---|
| 11TQM | Montana | Evergreen Forest (53%) Shrub/Scrub (22%) Grassland/Herbaceous (13%) | 59% | 39 | 226 (Landsat-9) |
| 11SPS | California | Shrub/Scrub (39%) Barren Land (29%) Cultivated Crops (21%) | 15% | 103 | 195 (Landsat-8) |
| 12TVL | Utah | Shrub/Scrub (42%) Deciduous Forest (19%) Evergreen Forest (13%) | 47% | 55 | 282 (Sentinel-2B) |
| 13SEC | Colorado | Grassland/Herbaceous (78%) Shrub/Scrub (8%) Evergreen Forest (4%) | 36% | 85 | 268 (Sentinel-2A) |
| 14TNP | South Dakota | Cultivated Crops (53%) Pasture/Hay (30%) Grassland/Herbaceous (7%) | 39% | 75 | 140 (Landsat-9) |
| 15SWB | Missouri | Deciduous Forest (51%) Pasture/Hay (39%) Developed, Open Space (4%) | 39% | 75 | 078 (Sentinel-2B) |
| 17RNL | Florida | Pasture/Hay (28%) Woody Wetlands (20%) Open Water (18%) | 49% | 63 | 019 (Landsat-9) |
| 18TWN | New York | Deciduous Forest (33%) Pasture/Hay (18%) Mixed Forest (15%) | 61% | 37 | 148 (Landsat-8) |
| 10TDM | Oregon / California | Evergreen Forest (53%) Shrub/Scrub (28%) Grassland/Herbaceous (9%) | 38% | 75 | 182 (Sentinel-2A) |

## 3. Method

### 3.1 HLS-GPT model structure

The HLS-GPT model uses generative pre-trained Transformers concepts developed for language modelling (Radford et al., 2018) and adopts a hierarchical architecture composed of three Transformer modules. Two Transformers are used to process the Landsat and Sentinel-2 HLS time series independently. This is because the Landsat and Sentinel-2 observations have different numbers of spectral bands (7 for Landsat and 11 for Sentinel-2). A third Transformer is used to integrate the concatenated sensor features based on an approach we developed for HLS-based within-season crop type classification (Zhang et al., 2025a). The two sensor-specific Transformers function similarly to "encoders" used in spatial–temporal deep learning frameworks for multi-source satellite data (Benson et al., 2024; Brown et al., 2025; Michel & Inglada, 2026), where plain neural networks or CNN "encoders" generate date-specific feature encodings but do *not* model interactions among different dates. In contrast, the HLS-GPT sensor-specific Transformers explicitly model temporal dependencies independently within the Landsat and within the Sentinel-2 time series, allowing meaningful feature representation even on dates with no observations before across-sensor feature vector concatenation. We found experimentally that this design outperforms the use of neural-network encoders without temporal interactions followed by concatenation of cross-sensor feature vectors with a single shared Transformer. This is because cross-sensor concatenation is performed along the Landsat and Sentinel-2 feature dimensions at each observation date, even though many dates have an observation from only one of the two sensors. Through asymmetric masking in the attention layer, each sensor-specific Transformer can still produce meaningful feature representations for dates without observations from that sensor before feature concatenation occurs. In this sense, the sensor-specific Transformers act as date-alignment networks.

The HLS time series were organized into a data structure to enable the two Transformers to separately encode the Landsat and Sentinel-2 sequences. For a given pixel location, let $T$ denote the maximum number of HLS good-quality acquisition days within a year ($T$=176 in this study see Fig. 2). This ensures that the fixed-length sequence can represent any 12-month time series with $T'$ good-quality HLS observations ($T' \leq T$ by definition), with the remaining $T - T'$ positions padded using fill values, which were set to −9999. In this study, $T'$ ranges from 8 to 176, with a mean of 44 and a median of 40, based on calendar-year 12-month periods. Note that during training, $T'$ can vary more broadly, as the model is trained on 12-month periods that are not restricted to calendar years. The time-series structure for a single pixel location is illustrated in Fig. 3 with $T$ Landsat and $T$ Sentinel-2 days for each of the 9 years. Gray cells denote fill values (i.e., –9999 for $T - T'$ positions) that occur when no good-quality observation is available for a sensor, and for the red-edge and broad NIR bands not sensed by Landsat. The fill values are not involved in the computation, they are not fed into the attention values, as shown below. Note that the day-of-year (DOY) values always retain meaningful entries (not fill values) even when there is no good-quality observation, which is to enable the model to reconstruct reflectance for any day. The data structure can be conceptualized as an array with dimensions 312,950×$L$×12, where 312,950 is the number of training and test pixel locations, $L$=2×$T$×9 represents the length of the nine-year time series (2 sensors × $T$ maximum number of annual observations × 9 years), and 12 corresponds to 11 reflectance band values and a single DOY value. Prior to being ingested into the model, all the HLS reflectance values were z-score normalized using the band specific mean and standard deviation values computed from the training dataset. The DOY values were linearly transformed over the range [0, 2] (see Section 3.2).

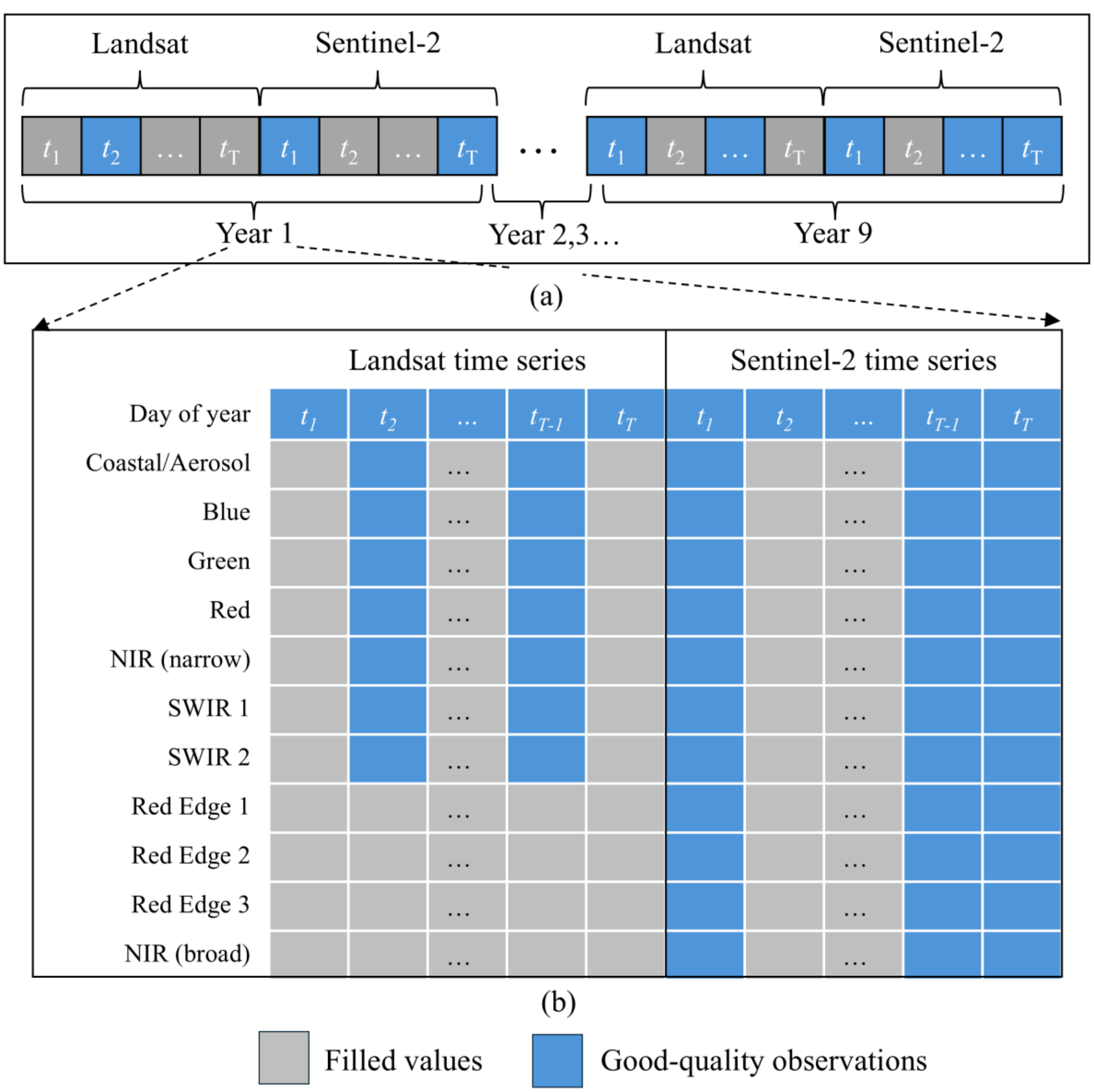

**Fig. 3.** HLS time series structure used in the study: a) the full nine-year HLS reflectance time series, and (b) detail of the first year. The blue cells represent good-quality observations and the gray cells indicate fill values. Note that this nine-year structure was used only for training, whereas the HLS-GPT model application used only a single 12-month time series.

The HLS-GPT architecture is illustrated in Fig. 4 and consists of four main components: (1) Input embedding. For each 12-month HLS single-pixel time series, a mask vector was constructed for the Landsat and Sentinel-2 bands at dates lacking good-quality observations, with the masked reflectance values set to 0. The 7-band Landsat and 11-band Sentinel-2 reflectance values were linearly projected into $d$-dimensional feature vectors for each day using two independent and trainable projectors, respectively, where $d$ is the Transformer hidden feature vector dimension. The DOY time series was linearly projected into a $d$-dimensional feature vector with trainable coefficients. In addition, two trainable $d$-dimensional feature vectors were assigned to represent the two sensor identities, 0 for Landsat and 1 for Sentinel-2, respectively. The three feature vectors derived from reflectance, DOY and sensor identity at each day were summed to derive a single feature vector for the day as the day feature representation before being fed into the Sensor-specific Transformers. (2) Sensor-specific Transformers. To account for differences between sensors, the model first captured temporal dependencies within each sensor data stream independently. Each sensor sequence was passed through $n_1$ identical Transformer blocks. Each block (Fig. 4b) consisted of a multi-head attention (MHA) layer, two layer-normalizations, and a feed-forward network (FFN) following the standard Transformer structure (Vaswani et al., 2017). Residual connections and dropout were applied after both the MHA and FFN modules. The MHA was the only layer capable of modeling temporal correlations, with queries, keys, and values derived from the input feature vectors. Attention weights were computed as the scaled dot product between queries and keys and applied to the values to produce a new feature vector that captured temporal dependencies. An MHA masking operation enforced that queries cannot attend to days of year with no good quality observations. The masking was asymmetric: queries from days of year without good-quality observations can still attend to days with good-quality observations for the reflectance reconstruction. As illustrated in Fig. 4c, consider three days: $t_1$ and $t_3$ with good-quality observations, and $t_2$ without a good-quality observation. Although $t_2$ contained no valuable spectral information, the input embedding still provided it with temporal and sensor identity information (otherwise the model would be unaware of which dates and sensor reflectance to reconstruct). The mask was applied to the $t_2$ key so that $q_1k_2 = 0$, $q_3k_2 = 0$ but $q_2k_1 \neq 0$, $q_2k_3 \neq 0$. Thus, queries at $t_1$ and $t_3$ would not attend to $t_2$, whereas a query at $t_2$ could leverage information from $t_1$ and $t_3$. (3) Cross-sensor fusion and Transformer. The outputs from the two sensor-specific Transformers were concatenated along the feature dimension, resulting in representations of size $T\times2d$ where $d$ is the hidden feature vector dimension of the sensor-specific Transformers. These concatenated features were then passed through additional Transformer blocks to facilitate cross-sensor information exchange. (4) Output projection. In the final stage, two parallel dense layers with sigmoid activation mapped the cross-sensor Transformer outputs back to the target spectral space of 11 bands. One dense layer produced the Landsat reconstruction, while the other produced the Sentinel-2 reconstruction. The two outputs were then concatenated to form the final time series reconstruction with dimensions ($2T\times11$) to store the $T\times11$ Sentinel-2 bands and the $T\times7$ Landsat bands (the other $T\times4$ is a dummy output). Note that the sigmoid activation was used in the last layer to enforce physically plausible spectral reflectance over a valid [0, 1] reflectance range.

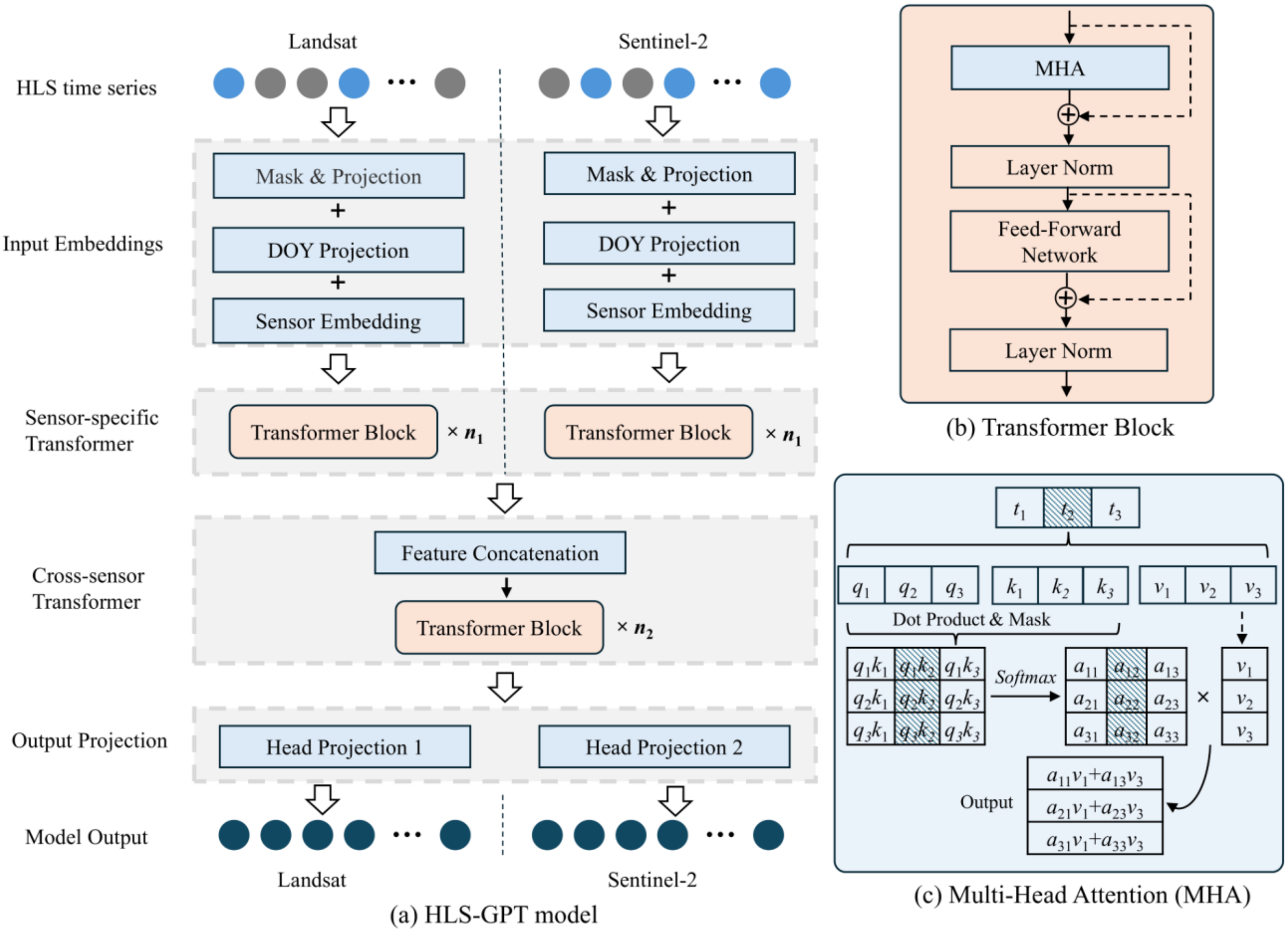

**Fig. 4.** The HLS-GPT model architecture that consists of four main components: input embeddings, sensor-specific Transformer, cross-sensor fusion and Transformer, and output projection, showing (a) Gray and blue solid circles in the input sequence that represent fill values and good-quality observations, respectively, (b) the Transformer block as the basis unit of the Transformer, (c) detail of the multi-head attention (MHA) module, with shaded squares to denote masked days of year, and showing three days of year ($t_1$, $t_2$, $t_3$) to illustrate the masking mechanism.

## 3.2 HLS-GPT training strategy

The HLS-GPT model was trained using all the CONUS training pixel data and has >17 million learnable coefficients. The HLS-GPT model coefficients were initialized in the conventional way with small random values falling in a [–0.02, 0.02] range. The training strategy is illustrated in Fig. 5 that we refer to, for brevity, as masked temporal reconstruction (MTR). The first step is to randomly crop 12-month reflectance subsequences from the 9-year time series, allowing each 12-month sequence to begin on an arbitrary date. Specifically, at each training pixel location, 12-month time periods were extracted with random start days from each of the first eight years (a start day in the eighth year will necessarily include observations from the ninth year). The start day was randomly selected for each pixel location, for each of the eight 12-month subsequences, and for each training epoch. This ensures that all observations across the 9-year time series have the potential to be sampled, and enables the trained HLS-GPT model to capture reflectance seasonality, phenological patterns, and inter-annual variations. To ensure sufficient temporal information to learn from, only 12-month reflectance subsequences with ≥8 HLS good-quality observations were considered in the training. As noted earlier (Section 2.2) 97.8% of the CONUS pixels locations had ≥8 good-quality observations.

In each 12-month reflectance subsequence, the date (i.e. DOY value) of each valid observation was normalized linearly as $((DOY-1)/366+(Year-Year_{min}))$, where $Year_{min}$ is the calendar year that the first observation in the subsequence fell within, providing DOY values stored in the range 0 - 2. The normalized DOY values were encoded using a learnable linear transformation. This approach was adopted after comparison with the commonly used sine–cosine positional encoding (Vaswani et al. 2017). In our case the learnable linear projection encoding worked more reliably and it has also been used in other remote sensing studies, for example, by Feng et al. (2025), Ravirathinam et al. (2025), and Wang et al. (2025c).

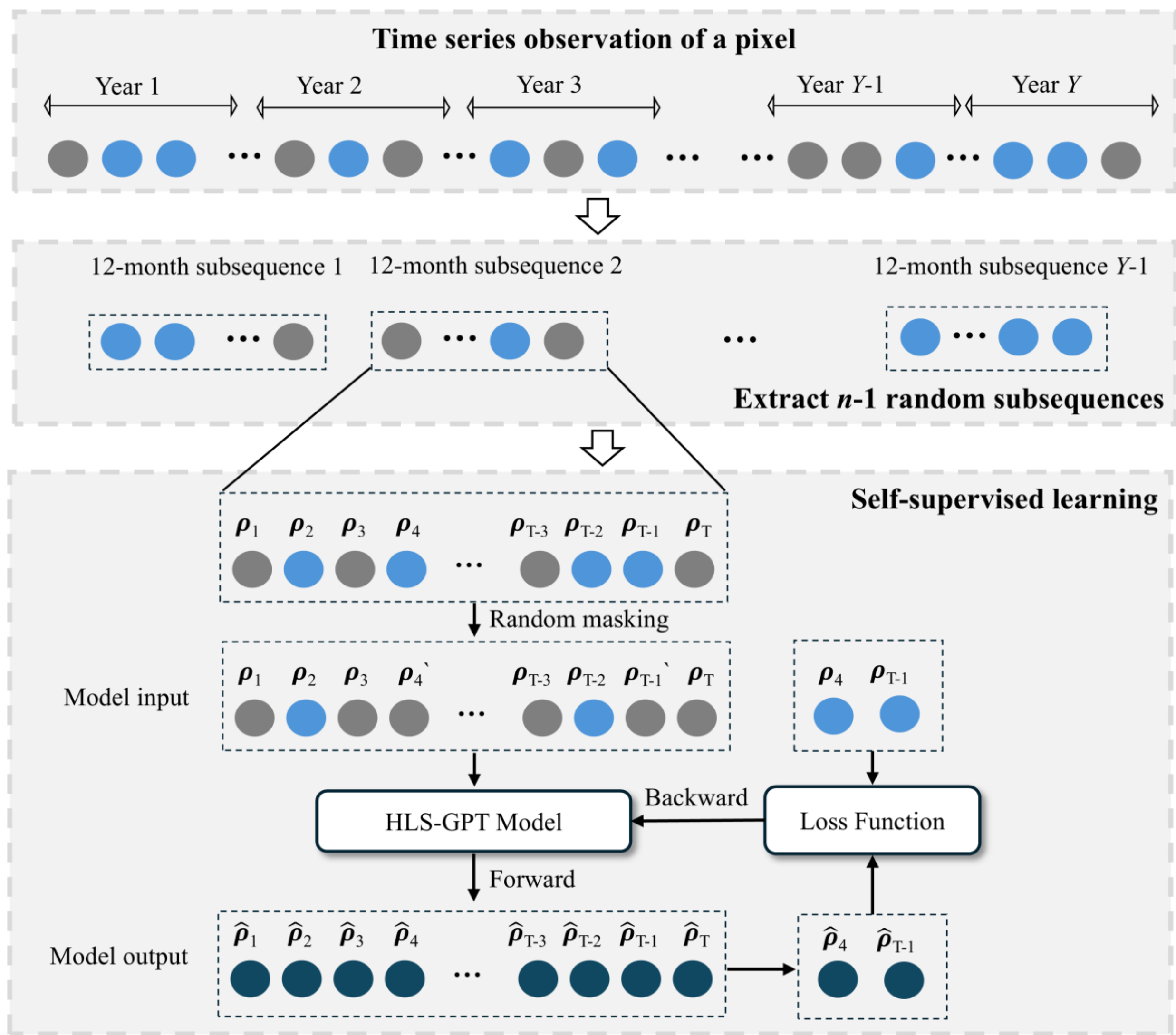

**Fig. 5.** The masked temporal reconstruction (MTR) training strategy used to train the HLS-GPT model. The blue circles represent good-quality observations, and the gray circles represent fill values (or masked good quality) values. A year 2 subsequence is used to illustrate the self-supervised learning process, where $\boldsymbol{\rho}_4$ and $\boldsymbol{\rho}_{T-1}$ are randomly masked good-quality reflectance observation vectors (11 bands) for loss computation against the model predicted reflectance vectors, $\widehat{\boldsymbol{\rho}}_4$ and $\widehat{\boldsymbol{\rho}}_{T-1}$. The parameters $T$ and $Y$ correspond to 176 days and 9 years, respectively, for the HLS results presented in this paper.

The $T'$ valid observations in each $T$-length subsequence were randomly separated into predictors ($T'/2$) and supervision targets ($T'/2$). The mask vector for each time series was updated to reflect this random masking operation, i.e., the new mask including the $T'/2$ predictor positions and $T$ - $T'$ positions with no valid observations. The self-supervised learning was undertaken iteratively over many training epochs. The core step was to randomly mask 50% of the good-quality observations in each 12-month subsequence for supervision. For example, in Fig. 5 the data on day $t_4$ (a vector of 11 reflectance values denoted $\boldsymbol{\rho}_4$) and on day $t_{T-1}$ (a vector of 11 reflectance values denoted $\boldsymbol{\rho}_{T-1}$) were masked and treated as targets ($y$) for supervision, while the remaining good-quality observations were used as predictors ($x$). The model predicted reflectance, i.e., a vector of 11 predicted reflectance values for days $t_4$ and $t_{T-1}$, denoted $\widehat{\boldsymbol{\rho}}_4$ and $\widehat{\boldsymbol{\rho}}_{T-1}$, were then compared with the corresponding observed reflectance, $\boldsymbol{\rho}_4$ and $\boldsymbol{\rho}_{T-1}$, and the loss was backpropagated to update the HLS-GPT model coefficients. A conventional RMSE loss function was used without any temporal smoothing-based loss such as that implemented by the SSLI (Liu et al., 2024). This is because smoothing could suppress abrupt changes, thereby limiting the HLS-GPT model's ability to capture rapid reflectance changes. Randomly masking 50% of the observations promotes temporally diverse reflectance sequences, because the number of available good-quality observations varies across 12-month periods and pixel locations. We also tested other masking proportions and strategies, such as randomly masking between 10% and 90% of the observations, but these either performed worse than or similarly to the fixed 50% masking strategy.

The HLS-GPT model was implemented in the TensorFlow framework, and all experiments were conducted on a high-performance computing platform equipped with two NVIDIA A100 GPUs (each 80 GB GPU memory). Following our previous work (Zhang et al., 2024; Liu et al., 2024; Zhang et al., 2025a), we performed hyperparameter tuning and experimental evaluation to determine the optimal configuration. The final settings were as follows: the embedding dimension in the input stage was set with $d$=256; the sensor-specific Transformer module contained three Transformer blocks ($n_1$=3); the cross-sensor fusion Transformer included four Transformer blocks ($n_2$=4); and the number of attention heads was set to $h$=8. Model optimization was performed using the AdamW optimizer with a weight decay of $1\times10^{-4}$. The learning rate was set to 0.0005, linearly warmed up to this value, and subsequently decayed following a cosine schedule. A batch size of 1024 was used, and the model was trained for a total of 30 epochs that were found to be sufficient to provide the results described in Section 4.

### 3.3 Qualitative visual assessment of single pixel reconstructed time series

Illustrative year 2023 reflectance time series were reconstructed at example pixel locations selected from the nine evaluation tiles. These included two deciduous forest pixels with markedly different numbers of good-quality observations, one alfalfa pixel that exhibited complex temporal variation due to multiple alfalfa harvests, one shrubland pixel that was burned and had abrupt reflectance changes, and an open water pixel. The Landsat and Sentinel-2 fill values in each pixel time series were reconstructed and the observed and reconstructed reflectance value were plotted as a function of DOY. Due to space constraints the red and NIR reflectance, and the derived normalized difference vegetation index (NDVI), only were illustrated as they are commonly used for vegetation monitoring (Huete et al., 2002) and are also useful for diagnosis of residual atmospheric correction errors, and undetected/sub-pixel cloud and shadows (Roy et al., 1997; Ju et al., 2010).

### 3.4 Quantitative HLS-GPT reconstruction evaluation

The HLS-GPT model reconstruction results were evaluated quantitatively considering the test data 9-year HLS good-quality single pixel reflectance time series. For each of the 62,590 test pixel locations, eight 12-month subsequences were generated following the masked temporal reconstruction strategy, resulting in 415,797 12-month subsequences containing ≥8 HLS good-quality observations. A random proportion of the good-quality observations in each subsequence were masked (following two different strategies described below), and then the reflectance time series from the remaining observations was reconstructed and compared with the masked good-quality observations.

The reconstruction error was quantified using the conventional band specific RMSE metric, and the relative RMSE that reduces the influence of different reflectance values among spectral bands, was also quantified as:

$$\mathrm{RMSE}_{b,sensor} = \sqrt{\frac{\sum_{i=1}^{n}\sum_{t=1}^{k_i}\left(\rho_{b,sensor,t}^{i} - \hat{\rho}_{b,sensor,t}^{i}\right)^2}{\sum_{i=1}^{n} k_i}} \quad (1)$$

$$rRMSE_{b,sensor} = \frac{\mathrm{RMSE}_{b,sensor}}{\frac{\sum_{i=1}^{n}\sum_{t=1}^{k_i}\rho_{b,sensor,t}^{i}}{\sum_{i=1}^{n} k_i}} \times 100\% \quad (2)$$

where $\mathrm{RMSE}_{b,sensor}$ is the reconstruction RMSE for spectral band *b* of *sensor* (Landsat or Sentinel-2), $rRMSE_{b,sensor}$ is the relative RMSE, *n* is the number of 12-month test subsequences (*n*= 415,797), $k_i$ denotes the number of good-quality HLS observations in subsequence $i$, $\rho_{b,sensor,t}^{i}$ is the observed good-quality HLS reflectance for *sensor* band *b* from subsequence *i* on day *t*, and $\hat{\rho}_{b,sesnor,t}^{i}$ is the corresponding HLS-GPT reconstructed reflectance.

In addition, the all-spectral-band RMSE was calculated as:

$$\text{RMSE}_{all_bands,sensor} = \sqrt{\frac{\sum_{b=1}^{B}\sum_{i=1}^{n}\sum_{t=1}^{k_i}(\rho_{b,sensor,t}^{i}-\hat{\rho}_{b,sensor,t}^{i})^2}{B*\sum_{i=1}^{n}k_i}} \quad (3)$$

where $\text{RMSE}_{all_bands,sensor}$ is the reconstruction RMSE for all $B$ sensor bands ($B$ = 7 for *sensor* = Landsat and $B$ = 11 for *sensor* = Sentinel-2) and the other symbols are defined as for Eqs. (1) and (2).

Two strategies were used to capture different reconstruction scenarios. The first strategy termed *leave-one-observation-out* is where each good-quality observation in a 12-month subsequence was masked and reconstructed using the remaining good quality observations. This was undertaken exhaustively and independently for all the good-quality observations in the subsequence. For example, if there were $k$ good-quality reflectance values in the 12-month subsequence, then there would be $k$ sets of single observation reconstructed reflectance values each reconstructed from $k$-1 values. The second strategy termed *random masking*, is where a fixed proportion (10%, 20%, …, 90%) of the good-quality observations in the 12-month subsequence were randomly masked and reconstructed from the remainder. For example, if there were $k$ good-quality reflectance values in the subsequence, then a fixed proportion ($p$%) of them would be masked and their reflectance values reconstructed from the remaining $(100 - p)$% good-quality observations. For both strategies the $\text{RMSE}_{b,sensor}$, $rRMSE_{b,sensor}$ and $\text{RMSE}_{all_bands,sensor}$ values were derived.

3.5 Tile based image reconstruction evaluation and comparison with other methods

3.5.1 Evaluation metrics

To demonstrate the spatial applicability of the HLS-GPT model, the reflectance for a single reconstruction target day in 2023 was reconstructed for all the 30 m pixels in each of the nine CONUS HLS evaluation tiles. Recall that the HLS data for these nine tiles were not used to train or test the HLS-GPT model (Fig. 1 red boundaries). The tile reconstruction target day (summarized in Table 1, last column) for each of the 9 tiles was selected by searching through all the HLS tile data acquired over each tile in 2023 and finding the acquisition day when >85% of the 3660 × 3660 tile pixels were observed by a sensor and all the observed pixels were good quality. The search was initiated from July 15th 2023 backwards and forwards in time, and in some tiles, such as the Florida and Utah tiles, the first date meeting these criteria was near the beginning or the end of the year (Table 1, last column). The selected reconstruction target days were sensed either by a Landsat (5 tiles) or Sentinel-2 (4 tiles) sensor. This mix of different reconstruction days and sensors is helpful to demonstrate the robustness of the HLS-GPT model reconstruction.

The reconstruction target day tile pixels were masked and the remaining good-quality observations (acquired between January 1st 2023 and December 31st 2023) were used to reconstruct the target day reflectance at each pixel location. The reconstruction error considering all bands for a specific tile was calculated as:

$$RMSE_{all_bands}^{tile} = \sqrt{\frac{\sum_{b=1}^{B}\sum_{i=1}^{npix_{tile}}\left(\rho_b^{tile,i} - \hat{\rho}_b^{tile,i}\right)^2}{B * npix_{tile}}} \quad (4)$$

where $RMSE_{all_bands}^{tile}$ is the reconstruction RMSE for a specific *tile* for all *B* bands sensed on the reconstruction target day, $npix_{tile}$ is the number of good quality 30 m tile pixel observations (typically close to 3660 × 3660), $\rho_b^{tile,i}$ is the observed good-quality HLS reflectance and $\hat{\rho}_b^{tile,i}$ is the corresponding reconstructed reflectance, *i* denotes the tile pixel location represented for convenience in Equation (3) flattened from a 3660 × 3600 tile pixel row and column coordinate array to a vector of length $npix_{tile}$.

In addition, the reconstruction error considering all bands and all tiles for a specific land cover class was calculated as:

$$RMSE_{all_bands,class}^{all_tiles} = \sqrt{\frac{\sum_{b=1}^{B}\sum_{tile=1}^{9}\sum_{i=1}^{npix_{tile}} m_{class}^{tile,i} * \left(\rho_b^{tile,i} - \hat{\rho}_b^{tile,i}\right)^2}{B * \sum_{tile=1}^{9}\sum_{i=1}^{npix_{tile}} m_{class}^{tile,i}}} \quad (5)$$

where $RMSE_{all_bands,class}^{all_tiles}$ is the reconstruction RMSE for a specific land cover *class* across all the nine tiles for all *B* bands sensed on the reconstruction target day, $m_{class}^{tile,i}$ is a binary indicator, equal to 1 if pixel *i* in the tile belongs to the NLCD-defined land cover *class* and 0 otherwise, and the other symbols are defined as for Eq. (3). The $RMSE_{all_bands,class}^{all_tiles}$ has eight values, each corresponding to one of the eight NLCD Level-1 classes (water, developed, barren, forest, shrubland, herbaceous, planted/cultivated, and wetlands) that are illustrated in Fig. 1.

The reconstruction error for a given band was also summarized for all tiles as:

$$RMSE_{b}^{all_tiles} = \sqrt{\frac{\sum_{tile=1}^{9}\sum_{i=1}^{npix_{tile}}\left(\rho_b^{tile,i} - \hat{\rho}_b^{tile,i}\right)^2}{\sum_{tile=1}^{9} npix_{tile}}} \quad (6)$$

where $RMSE_{b}^{all_tiles}$ is the RMSE for spectral band *b* considering all nine tiles together, $npix_{tile}$ is the number of good quality 30 m tile pixels for a specific *tile*, $\rho_b^{tile,i}$ is the observed good-quality

HLS reflectance and $\hat{\rho}_b^{tile,i}$ is the corresponding reconstructed reflectance, $i$ denotes the tile pixel location represented for convenience in Equation (6) flattened from a 3660 × 3600 tile pixel row and column coordinate array to a vector of length $npix_{tile}$.

The structural similarity index (SSIM) (Wang et al., 2004) was used to quantify the ability of the HLS-GPT model to reconstruct spatial information reliably across each of the 9 tiles. This is pertinent as the HLS-GPT model reconstructs from single-pixel time series information only. The SSIM metric accounts for spatial structure similarity across an image, and is bounded from 0 to 1, with values closer to 1 indicating higher structural similarity. The SSIM was calculated for each sensor band and for each tile independently ($SSIM_b^{tile}$) using the default index calculation settings described in (Wang et al., 2004). To report overall spatial reconstruction accuracy, the $SSIM_b^{tile}$ was averaged across all bands to obtain a single value ($SSIM_{all_bands}^{tile}$) and was also averaged for each band across all tiles to obtain a single value ($SSIM_b^{all_tiles}$). Note that land cover class-specific SSIM values were not derived because frequent class boundaries and isolated class pixels prevent meaningful SSIM computation.

### 3.5.2 Comparison with "benchmark" methods

At the tile level, we compared the HLS-GPT performance with conventional time series fitting methods, namely with HANTS (Zhou et al., 2015) and the Savitzky–Golay filter (Chen et al., 2004; Chen et al., 2021; hereafter referred to as SG filter), and with the recent NASA–IBM Prithvi-EO-2.0 deep learning foundation model (Godwin et al., 2025; Szwarcman et al., 2025; hereafter referred to as Prithvi). Among the open-source, national-scale deep learning models summarized in Appendix A, U-TILISE (Stucker et al., 2023) and GANFilling (Gonzalez-Calabuig et al., 2025) were not included in the comparison because they were trained only over limited regions in Europe and support reconstruction of only a subset of Sentinel-2 bands (i.e., red, green, blue and NIR), making them not directly comparable to the HLS-GPT model.

The HANTS and SG conventional time series fitting methods require no training data and can provide robust and interpretable results for noisy and irregular observations. HANTS is a harmonic model that assumes periodicity and fits the entire time series using a set of sinusoidal functions (Zhou et al., 2015). The SG filter smooths time series by fitting low-order polynomials within a moving window, preserving local features such as peaks and trends while reducing noise (Chen et al., 2021). The input parameters for HANTS and SG filter were set following their original implementations. As the reconstruction target in this study is surface reflectance rather than NDVI, the lower threshold used in HANTS was set to 0.

The Prithvi is based on a ViT-H (huge Vision Transformer) architecture and was trained using the MAE strategy with >1 million globally distributed sequences of four <20% cloudy HLS 256×256 30 m pixel patches acquired within a two-year period (Jakubik et al., 2023; Szwarcman et al., 2025). The model accepts four HLS 224×224 30 m pixel patches containing six HLS bands (blue, green, red, narrow NIR, SWIR1, and SWIR2) randomly spatially cropped from the input 256×256 patches. Following the Prithvi recommended implementation guidelines (Szwarcman et al., 2025), we used the nearest three dates with >80% good quality observations to the Year 2023 reconstruction target day (Table 1, last column) (the three selected dates are provided in Appendix B). The reconstruction target day HLS tile data were masked, and the four HLS tile data sets (three selected and the masked target day) were fed into the Prithvi model to reconstruct the tile data on the target day with a stride of 224 pixels to reconstruct the data across the tile. We selected the Prithvi-EO-2.0-600M-TL variant, as it provides the highest performance and the largest number of learnable parameters (~600 million) among the available models (Szwarcman et al., 2025). The pretrained model weights are publicly available at: https://huggingface.co/ibm-nasa-geospatial/Prithvi-EO-2.0-600M-TL

The reconstruction $RMSE_{all_bands}^{tile}$ , $RMSE_{b}^{all_tiles}$, and SSIM metrics were calculated for HANTS, SG, Prithvi, and the HLS-GPT reconstruction results considering the same six HLS bands used to train the Prithvi model. The same HLS-GPT model described in Section 3.3 was used to reconstruct 7-band Landsat or 11-band Sentinel-2 reflectance depending on the tile (Table 1), but only six common bands were selected for comparison with the Prithvi model. To ensure a fair comparison, these metric calculations excluded any tile pixel locations where cloud was labelled or there were no observations (missing data) in any of the four HLS tile data sets used to derive the Prithvi reconstruction results. This is because these pixel locations negatively impact the Prithvi reconstruction results (illustrated in Section 4.3).

## 4. Results

4.1 Illustrative single pixel HLS-GPT reconstructed time series

Figs. 6 to 10 show, for different example pixel locations, the year 2023 red and NIR good-quality HLS observation reflectance (black) and the reconstructed reflectance on days without an HLS Landsat and/or Sentinel-2 observation (orange) that were reconstructed from the observed values. In addition, the NDVI derived from the observed and reconstructed red and NIR reflectance is shown. The Fig. 6 results are for a deciduous forest pixel that had a relatively high number of good-quality HLS observations (22 Landsat and 65 Sentinel-2). The reconstructed red, NIR reflectance, and NDVI time series (orange) show strong agreement with the observed values (black), effectively capturing deciduous forest growth and senescence. The small HLS reflectance differences between the observed Landsat (black circles) and Sentinel-2 (black triangles) are also evident in the reconstructed Landsat (orange circles) and Sentinel-2 (orange triangles) reflectance time series.

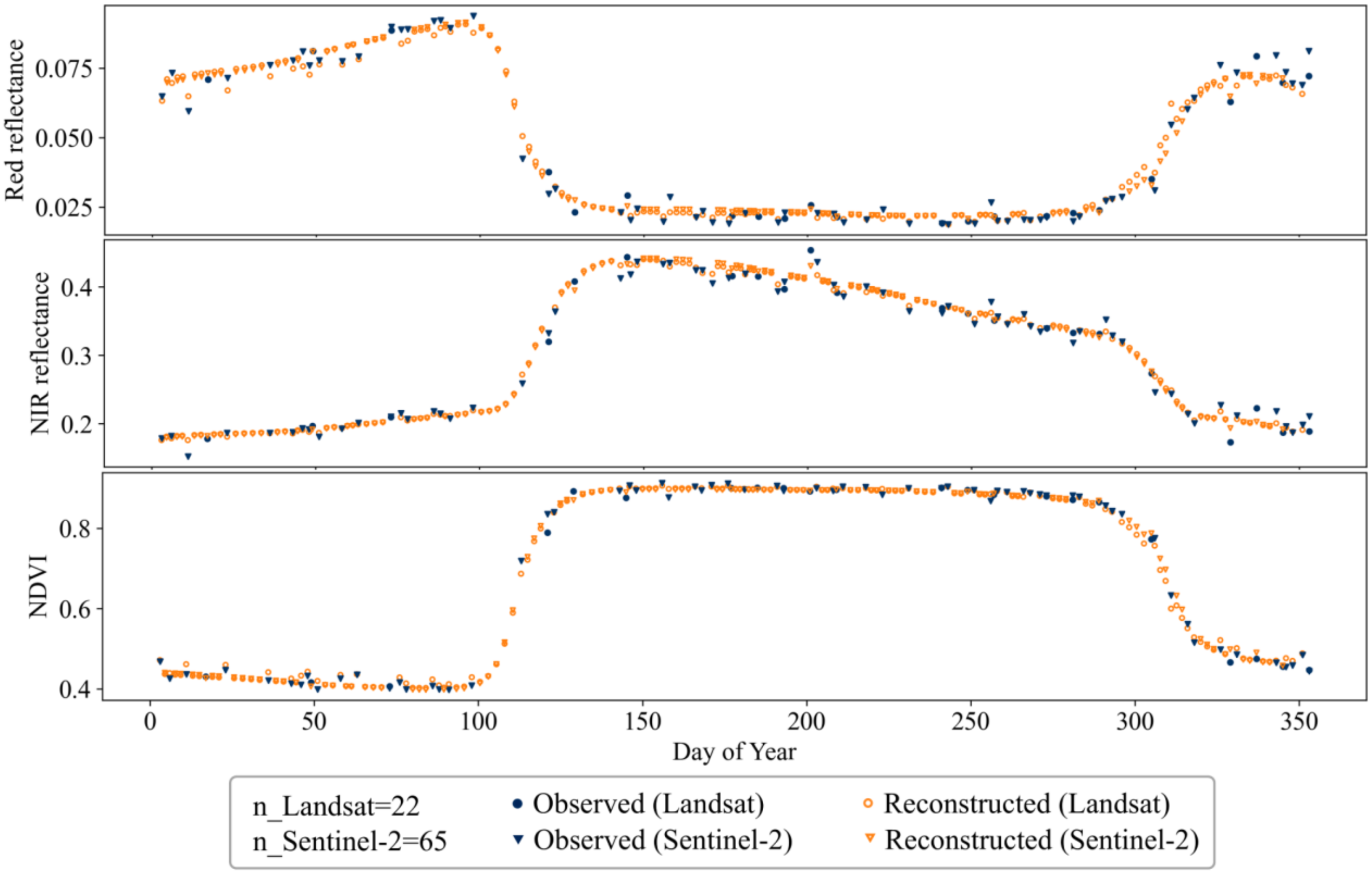

**Fig. 6.** Red, NIR and NDVI year 2023 reconstructed (orange) values for a deciduous forest pixel located in Missouri HLS tile 15SWB (Table 1) acquired at 92.9333°W, 36.9669°N. The n_Landsat and n_Sentinel-2 values denote the number of good-quality HLS processed Landsat and Sentinel-2 observations (shown black) in 2023.

Fig. 7 shows time series for another deciduous forest pixel, located in New York state, that had a relatively small number of good-quality HLS observations (9 Landsat and 10 Sentinel-2). Even under these sparse observation conditions, the HLS-GPT model reconstructed plausible deciduous forest reflectance phenological dynamics. The reconstructed reflectance difference between Landsat and Sentinel-2 sensors is most apparent in the red band with a mean reconstruction difference between sensors of 0.0023 and a mean relative difference 5.07% which is similar to the typical 4% mean relative HLS processed sensor difference reported by Ju et al. (2025).

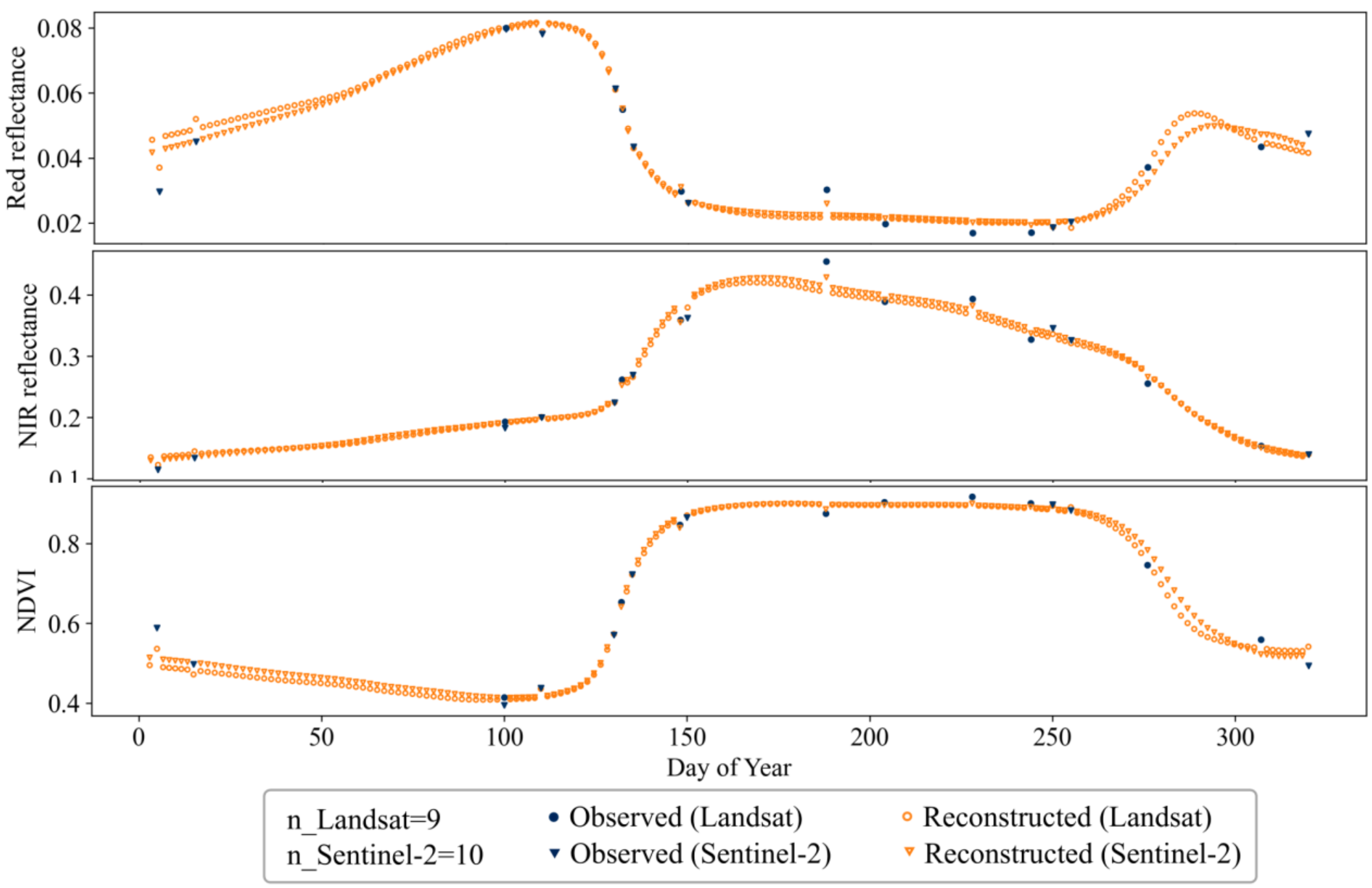

**Fig. 7.** Red, NIR and NDVI year 2023 reconstructed (orange) values for a deciduous forest pixel located in New York state HLS tile 18TWN (Table 1) acquired at 74.5253°W, 42.5143°N. The n_Landsat and n_Sentinel-2 values denote the number of good-quality HLS processed Landsat and Sentinel-2 observations (shown black) in 2023.

Fig. 8 shows a time series for a Southern California alfalfa pixel that had a large number of good-quality Landsat (38) and Sentinel-2 (50) observations due to low cloud cover and lateral orbit swath overlaps (e.g., see Egorov et al., 2019). Benefiting from a long growing season, alfalfa in Southern California is harvested multiple times per year, and the reconstructed reflectance time series reveal eight "growth–harvest–regrowth" cycles. During the alfalfa growth stages, increasing chlorophyll absorption reduced the red reflectance below 0.05, while NIR reflectance and NDVI rose rapidly to peak values (NDVI ≈ 0.8, NIR between 0.4–0.6). The HLS-GPT reconstructed reflectance and derived NDVI captures the alfalfa temporal dynamics with a fidelity that conventional harmonic time series fitting (Roy and Yan, 2020) and even multi-sensor approaches (Houborg and McCabe, 2018) are unable to reliably reproduce.

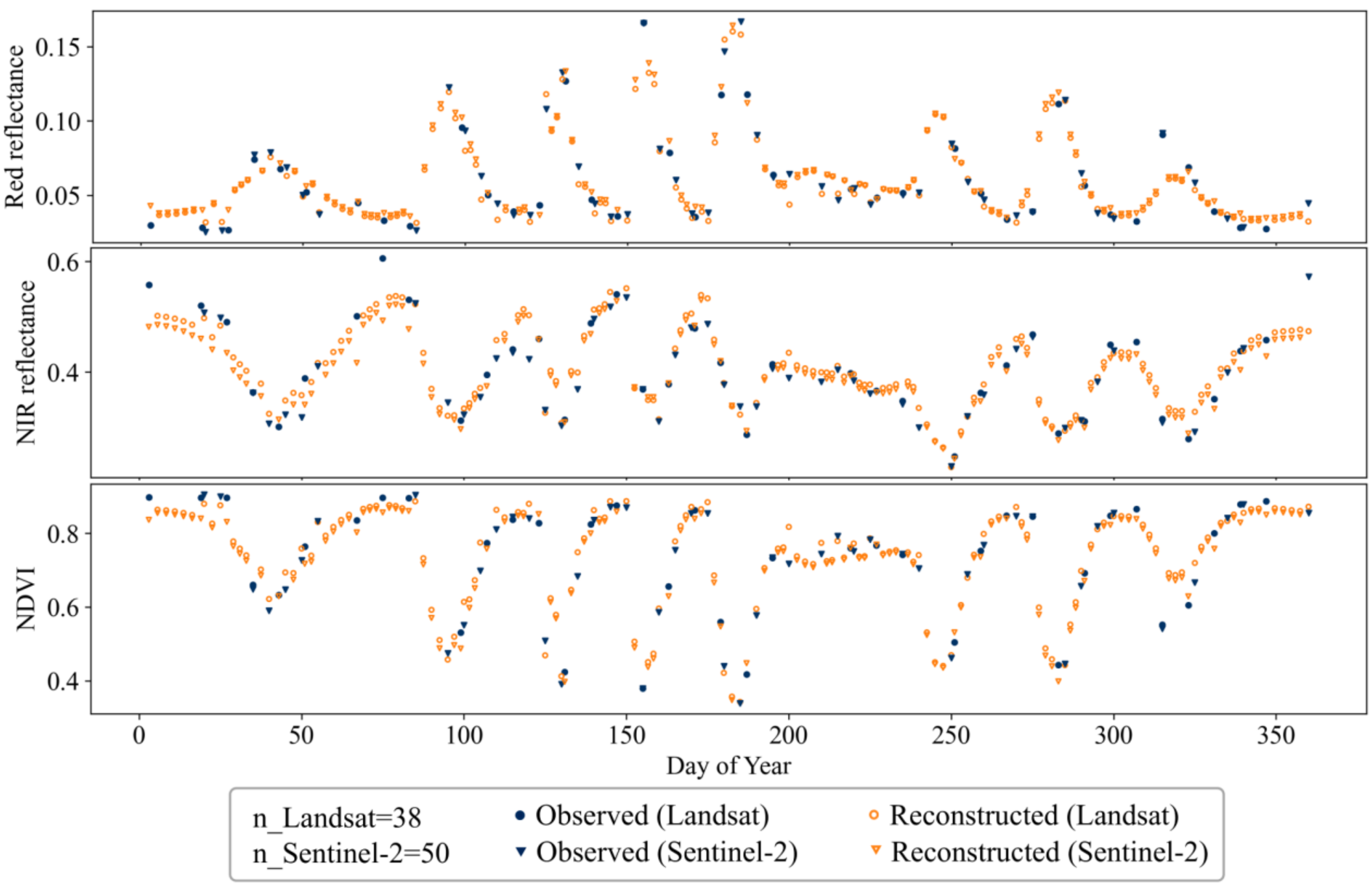

**Fig. 8.** Red, NIR and NDVI year 2023 reconstructed (orange) values for an alfalfa pixel located in California HLS tile 11SPS (Table 1) acquired at 115.4447°W, 32.9135°N. The n_Landsat and n_Sentinel-2 values denote the number of good-quality HLS processed Landsat and Sentinel-2 observations (shown black) in 2023.

Fig. 9 illustrates a time series for a shrubland pixel in Northern California that includes the Smith River Complex wildfire that began to burn August 15th 2023 (DOY = 227) (USFS, 2023). In general, fire reduces reflectance in most reflective wavelength bands and particularly in the NIR and may not substantially reduce the NDVI (Roy et al., 2002; Huang et al., 2016). In this example the observed NIR and also NDVI exhibited pronounced drops around DOY 230 that were captured in the reconstructed data. This demonstrates that the HLS-GPT model is able to reconstruct abrupt changes although the reconstruction smoothed the likely abrupt reflectance drop, which may be partially related to the temporal spacing of observations immediately before and after the fire started.

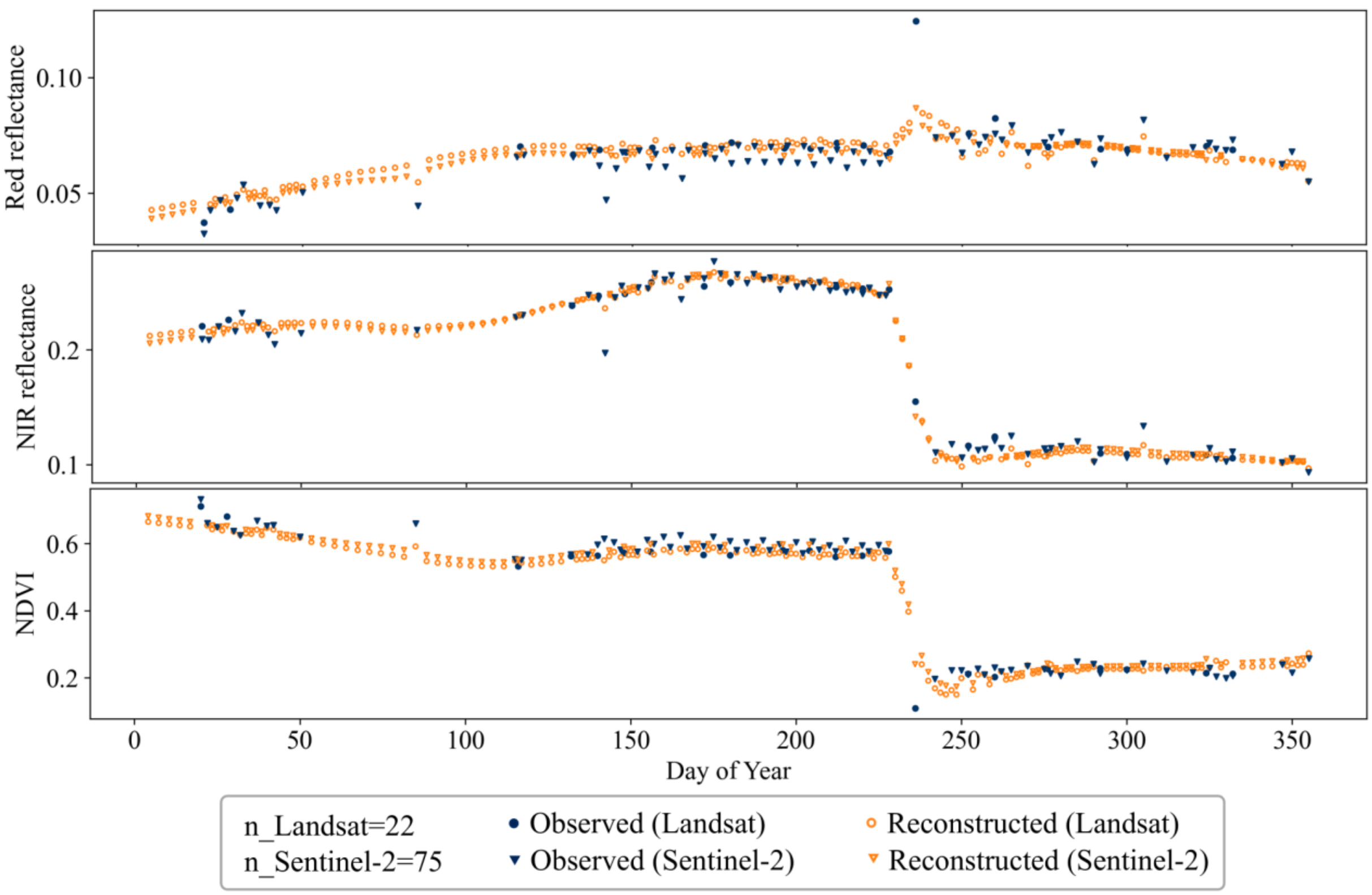

**Fig. 9**. Red, NIR and NDVI year 2023 reconstructed (orange) values for a burned shrubland pixel located in California HLS tile 10TDM (Table 1) at 124.0033°W, 41.9405°N. The n_Landsat and n_Sentinel-2 values denote the number of good-quality HLS processed Landsat and Sentinel-2 observations (shown black) in 2023.

Fig. 10 shows a time series for an open-water pixel in Lake Okeechobee, Florida. The observed red and NIR reflectance time series is noisy with abnormally high and also some low values that are due mainly to HLS cloud detection omission errors and residual atmospheric correction errors. Over the year, changes in the lake constituents, including suspended sediments and algae, and whitecaps on waves, may also influence the observed reflectance. Atmospheric correction over water is often unreliable due mainly to the low signal to noise of water bodies (Warren et al., 2019), and notably the LaSRC atmospheric correction algorithm used in the HLS processing was not optimized to correct reflectance over water (Crawford et al., 2023). Nominally, clouds reduce NDVI over vegetated surfaces (Roy, 1997; Ju et al., 2010) but over water the NDVI ratio formulation (the NIR minus red reflectance divided by their sum) causes large NDVI fluctuations due to the near-zero red and NIR reflectance values. This is apparent in the observed derived NDVI time series. Residual uncorrected atmospheric effects, particularly due to atmospheric aerosols, tend to increase visible reflectance over low reflectance surfaces (Roy et al., 2014b) and generally reduces NDVI over land (Van Leeuwen et al., 2006). Notably, the HLS-GPT reconstructed reflectance produced temporally smoother reflectance compared to the observed data with less temporal variation, and plausibly low red and NIR reflectance and derived NDVI close to zero throughout the year, that reflect the expected spectral characteristics of open-water surfaces. The reconstructed reflectance difference between the Landsat and Sentinel-2 sensors is most apparent in the middle of the year and is similar to the observed reflectance difference between sensors.

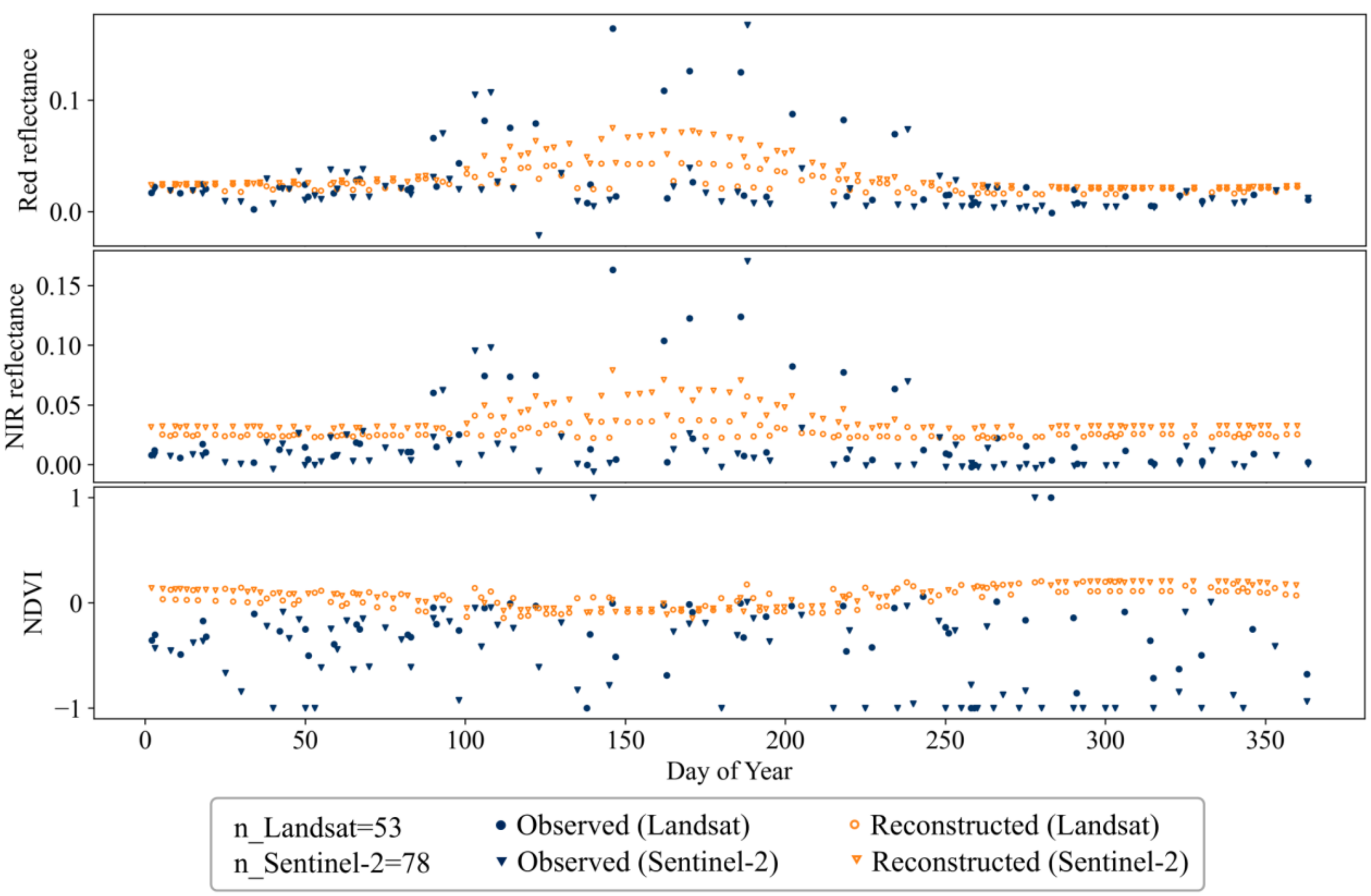

**Fig. 10.** Red, NIR and NDVI year 2023 reconstructed (orange) values for an open water pixel located in Lake Okeechobee in the Florida HLS tile 17RNL (Table 1) acquired at 80.8987°W, 27.0345°N. The n_Landsat and n_Sentinel-2 values denote the number of good-quality HLS processed Landsat and Sentinel-2 observations (shown black) in 2023.

## 4.2 HLS-GPT test data reconstruction evaluation

### 4.2.1 Leave-one-observation-out evaluation

In the leave-one-observation-out evaluation a single good-quality observation in a 12-month test subsequence was masked and reconstructed using the remaining good quality observations. This was undertaken exhaustively and independently for all the good-quality observations in each subsequence and for all 415,797 test subsequences. Table 2 summarizes the $\text{RMSE}_{b,sensor}$ (Eq. 1), $rRMSE_{b,sensor}$ (Eq. 2) and $\text{RMSE}_{all_bands,sensor}$ (Eq. 3) reconstruction values. The $\text{RMSE}_{b,sensor}$ values are low for the seven bands common to both the Landsat and Sentinel-2 sensors (visible, NIR and SWIR) varying from 0.014 to 0.026. For the Sentinel-2 only red-edge and broad NIR bands, the $\text{RMSE}_{b,sensor}$ values are comparable, varying from 0.017 to 0.022, indicating stable reconstruction performance even though there were no equivalent Landsat bands used as input for the reconstruction. The $\text{RMSE}_{all_bands,sensor}$ values considering all 7 Landsat and all 11 Sentinel-2 bands are 0.019 and 0.020, respectively, indicating similar reconstruction performance between sensors.

The $rRMSE_{b,sensor}$ values summarized in Table 2 reveal more pronounced spectral differences than the $\text{RMSE}_{b,sensor}$ values. This is expected as the $rRMSE_{b,sensor}$ formulae normalizes for the reflectance magnitude that can vary significantly spectrally (for example, vegetation has typically has low red reflectance but high NIR reflectance (e.g., Fig. 6)), and so enables more meaningful comparison among bands. The $rRMSE_{b,sensor}$ values decrease with wavelength from 34.0% (coastal blue) to 9.3% (NIR). This is likely because of the much greater uncertainty in atmospheric correction at shorter wavelengths over land and water surfaces (Ju et al., 2012; Roy et al., 2014b; Warren et al., 2019). The SWIR band $rRMSE_{b,sensor}$ values for both sensors vary from 9.8% to 12.8% and vary from 9.5% to 12.7% for the Sentinel-2 only red-edge and broad NIR bands. These results are discussed further in Section 5.2.

**Table 2**. The leave-one-observation-out evaluation $\text{RMSE}_{b,sensor}$ (Eq. 1), $rRMSE_{b,sensor}$ (Eq. 2) and $\text{RMSE}_{all_bands,sensor}$ (Eq. 3) results derived considering 415,797 12-month subsequences at 62,590 test pixel locations. The central wavelengths for the Sentinel-2 bands are shown for convenience (note that the 7 Landsat bands have slightly different spectral band passes and central wavelengths but these differences are minimized in the HLS processing).

| Band (central wavelength) | $\text{RMSE}_{b,sensor}$ | | $rRMSE_{b,sensor}$ | |
|---|---|---|---|---|
| | Landsat | Sentinel-2 | Landsat | Sentinel-2 |
| Coastal blue (0.443 μm) | 0.0147 | 0.0152 | 30.52% | 34.03% |
| Blue (0.490 μm) | 0.0146 | 0.0142 | 24.36% | 26.19% |
| Green (0.560 μm) | 0.0150 | 0.0151 | 16.38% | 18.33% |
| Red (0.665 μm) | 0.0163 | 0.0164 | 15.61% | 16.25% |
| NIR (narrow) (0.865 μm) | 0.0228 | 0.0237 | 8.98% | 9.29% |
| SWIR 1 (1.610 μm) | 0.0240 | 0.0259 | 9.76% | 10.06% |
| SWIR 2 (2.190 μm) | 0.0206 | 0.0218 | 11.84% | 12.78% |
| Red edge 1 (0.705 μm) | | 0.0173 | | 12.75% |
| Red edge 2 (0.740 μm) | | 0.0202 | | 10.03% |
| Red edge 3 (0.783 μm) | | 0.0221 | | 9.54% |
| NIR (broad) (0.842 μm) | | 0.0225 | | 9.49% |
| | $\text{RMSE}_{all_bands,sensor}$ | | | |
| All bands | 0.0187 | 0.0199 | NA | NA |

4.2.2 Random masking evaluation

In the random masking evaluation a fixed proportion (10%, 20%, …, 90%) of the good-quality test observations in each 12-month test subsequence were randomly masked and then reconstructed from the remaining observations. This is a more rigorous evaluation than the leave-one-observation-out approach as more observation data are typically masked prior to the reconstruction. Fig. 11 summarizes the $\text{RMSE}_{b,sensor}$ (Eq. 1) and $rRMSE_{b,sensor}$ (Eq. 2) values that, as expected, increase monotonically when a greater proportion of observations are masked. Despite this, the values are relatively similar for different masking proportions when 20%, 30%, 40%, or 50% of the good-quality test observations were randomly masked, indicating that the HLS-GPT model can reconstruct reflectance time series quite robustly even under sparse observation conditions as also evident in Fig. 7. The errors have similar magnitude to the leave-one-observation-out evaluation errors (Table 2) when 10% of the observations were masked. This is because the number of remaining observations for reconstruction is comparable between the two settings—for example, with an average of 69 observations in a 12-month subsequence, 68 remain in the leave-one-out evaluation and about 61 remain when 10% of the observations are masked. As before (Table 2) the reconstruction errors are greater for the shorter wavelength bands (i.e., visible bands) and are smaller for the non-visible red edge, NIR and SWIR bands.

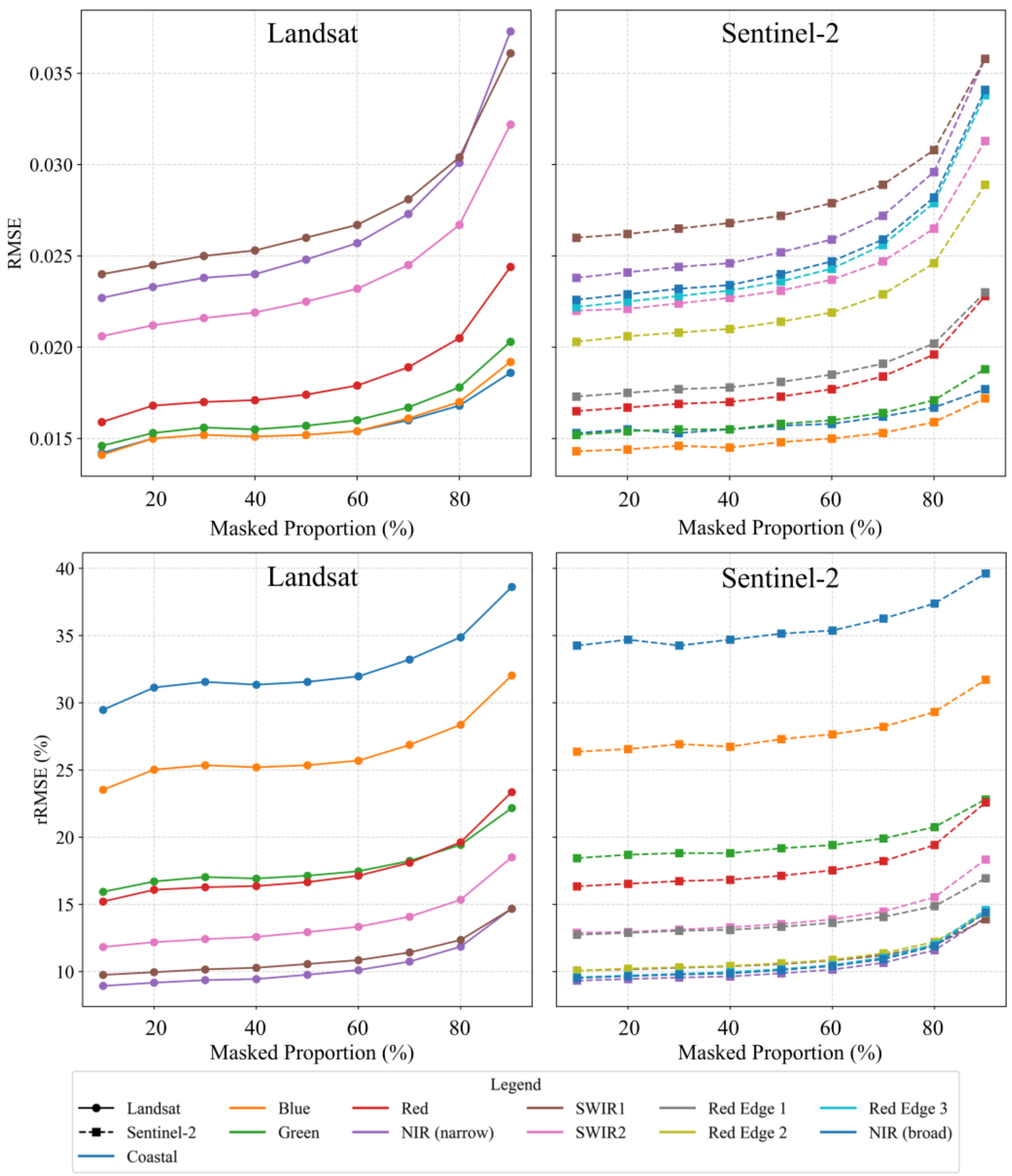

**Fig. 11.** The random masking evaluation $\text{RMSE}_{b,sensor}$ (Eq. 1) and $rRMSE_{b,sensor}$ (Eq. 2) results derived considering 415,797 12-month subsequences at 62,590 test pixel locations, found masking different proportions (10%, 20%, …, 90%).

4.3 Tile based image reconstruction evaluation and comparison with "benchmark" methods

Table 3 summarizes the $RMSE_{all_bands}^{tile}$ and $SSIM_{all_bands}^{tile}$ HLS-GPT and benchmark method reconstruction results for each of the nine evaluation tiles (the values shown in bold indicate higher reconstruction performance i.e., lower $RMSE_{all_bands}^{tile}$ values or $SSIM_{all_bands}^{tile}$ closer to 1). The HLS-GPT model outperformed HANTS, SG and Prithvi for all nine tiles and for both the spectral and structural reconstruction metrics. The HLS-GPT $RMSE_{all_bands}^{tile}$ values (ranging from 0.0052 to 0.0158) were consistently lower than those of Prithvi (ranging from 0.0089 to 0.0246), HANTS (ranging from 0.0079 to 0.0424) and SG (ranging from 0.0102 to 0.0321). Similarly the HLS-GPT SSIM values were consistently higher values across all tiles than the other methods.

**Table 3.** Tile specific $RMSE_{all_bands}^{tile}$ (Eq. 3) and $SSIM_{all_bands}^{tile}$ values for the HLS-GPT and benchmark method reconstructions (derived only for the six Landsat and Sentinel-2 bands used by Prithvi) for each of the 9 independent evaluation HLS tile reconstruction dates (Table 1). The bolded values indicate higher tile reconstruction performance (lower $RMSE_{all_bands}^{tile}$ or $SSIM_{all_bands}^{tile}$ closer to 1).

| Tile | $RMSE_{all_bands}^{tile}$ | | | | $SSIM_{all_bands}^{tile}$ | | | |
|---|---|---|---|---|---|---|---|---|
| | HLS-GPT | Prithvi | HANTS | SG | HLS-GPT | Prithvi | HANTS | SG |
| Oregon / California (10TDM) | **0.0052** | 0.0096 | 0.0079 | 0.0116 | **0.9950** | 0.9716 | 0.9926 | 0.9866 |
| Montana (11TQM) | **0.0062** | 0.0101 | 0.0127 | 0.0117 | **0.9875** | 0.9676 | 0.8521 | 0.9363 |
| California (11SPS) | **0.0139** | 0.0186 | 0.0297 | 0.0256 | **0.9566** | 0.9550 | 0.9460 | 0.9584 |
| Utah (12TVL) | **0.0103** | 0.0142 | 0.0388 | 0.0321 | **0.9772** | 0.9601 | 0.9371 | 0.9555 |
| Colorado (13SEC) | **0.0052** | 0.0086 | 0.0090 | 0.0102 | **0.9947** | 0.9783 | 0.9916 | 0.9882 |
| South Dakota (14TNP) | **0.0158** | 0.0246 | 0.0293 | 0.0241 | **0.9677** | 0.9229 | 0.9086 | 0.9476 |
| Missouri (15SWB) | **0.0086** | 0.0089 | 0.0095 | 0.0126 | **0.9801** | 0.9693 | 0.9746 | 0.9723 |
| Florida (17RNL) | **0.0114** | 0.0138 | 0.0424 | 0.0206 | **0.9586** | 0.9474 | 0.8015 | 0.9451 |
| New York (18TWN) | **0.0082** | 0.0165 | 0.0177 | 0.0185 | **0.9816** | 0.9473 | 0.7930 | 0.9466 |

As shown in Table 3, the conventional HANTS and SG (reconstruction methods generally achieve satisfactory performance over the Oregon/California, Colorado, Montana, and Missouri tiles. These tiles are dominated by relatively homogeneous vegetated surfaces, such as forest, shrubland, and grass/pasture (Table 1). These land cover types exhibit regular seasonal patterns that align well with the underlying assumptions of the time series fitting methods, which require smoothly changing temporal dynamics. However, their performance degrades substantially over more complex and heterogeneous landscapes, including cropland-dominated regions (e.g., the South Dakota tile) and wetland areas (e.g., the Florida tile), where reflectance dynamics are highly variable and non-stationary.

In contrast, the deep learning-based approaches (HLS-GPT and Prithvi) demonstrate more robust and consistent performance across diverse land cover types, indicating stronger generalization capability. In the following, we provide a detailed comparison between HLS-GPT and Prithvi across different tiles. The HLS-GPT model had particularly higher performance than Prithvi for the South Dakota, New York, and California tiles, where temporal dynamics and surface reflectance changes are complicated due to agricultural crop growth and harvesting. Fig. 12 shows the reconstructed red, green, and blue tile reflectance for these tiles, and also for the Missouri tile that is illustrated because it is dominated by non-crop vegetation and is predominantly deciduous forest and pasture/hay (Table 1). The reflectance observed on the target reconstruction day (left column) is shown for comparison with the reconstructed HLS-GPT (middle column) and Prithvi (right column) reflectance. The South Dakota tile reconstructed HLS-GPT reflectance is more similar to the observed reflectance (acquired in early Spring on DOY 140) than the Prithvi reconstructed reflectance that shows less apparent green vegetation. The clouds in the Prithvi reconstruction are due to cloud contamination in one of the three Prithvi input HLS tiles, and the missing pixels apparent in the north-west corner are due to missing observations in one of the three input HLS tiles. As noted in section 4.3, these pixels were not included in the RMSE and SSIM metrics calculations for either reconstruction method. The New York tile HLS-GPT reconstructed reflectance was also more similar to the observed reflectance (acquired in the Spring on DOY 148) than the Prithvi reconstruction which contains regions of noticeably browner vegetation particularly evident in a wetland area to the west of the Great Sacandaga Lake (the largest water body in the tile). The California tile contains the large Salton Sea water body in the north-west tile corner and an extensive region of irrigated agriculture to the south surrounded by desert. The observed reflectance was acquired in the Summer (DOY 195) and over land appears similar to both the reconstruction method results but not over the Salton Sea where both methods have different appearances to the observed reflectance. The Missouri tile observed reflectance was acquired in the Winter/early Spring (DOY 78) and at this scale there is not much apparent different although the Prithvi reconstructed tile reflectance appears slightly less reddish and greener than the HLS-GPT reconstructed and observed reflectance. These qualitive examples illustrate the efficacy of the HLS-GPT and Prithvi reconstruction approaches and the superiority of the HLS-GPT model.

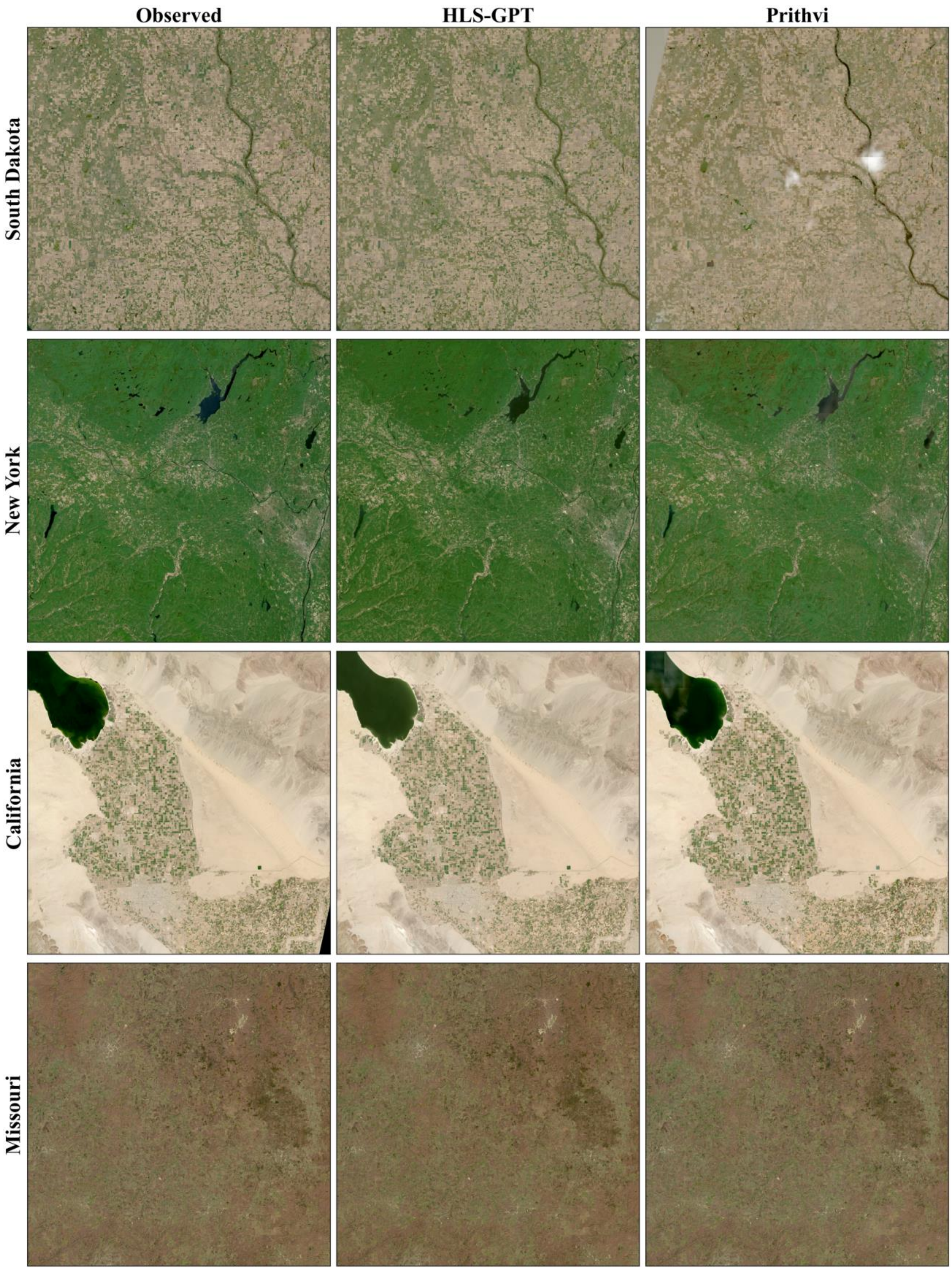
Observed
HLS-GPT
Prithvi
South Dakota
New York
California
Missouri

**Fig. 12.** True color (red, green, blue) reflectance for the South Dakota, New York, California, and Missouri independent evaluation tiles (rows). The left column shows the observed reflectance acquired on the reconstruction target date day of year (Table 1), the middle column shows the HLS-GPT reconstruction, and the right column shows the Prithvi reconstruction. Each tile is composed of 3660 × 3660 30 m pixels. The results for each tile (rows) are shown with the same contrast stretch to enable meaningful comparison.

Fig. 13 illustrates detailed 190 × 165 30 m pixel spatial subsets of the Fig. 12 California and New York tile results. The red boxes highlight regions where there are very apparent differences between the two reconstruction methods. Prithvi incorrectly reconstructed harvested fields as actively growing crops even though the nearest high-quality HLS tile data used by Prithvi were acquired only 5 days before and 5 days after the California reconstruction date and 13 days before and 2 days after the New York reconstruction date (Appendix B). In contrast, the HLS-GPT model was able to reconstruct the reflectance changes as also illustrated in Fig. 9.

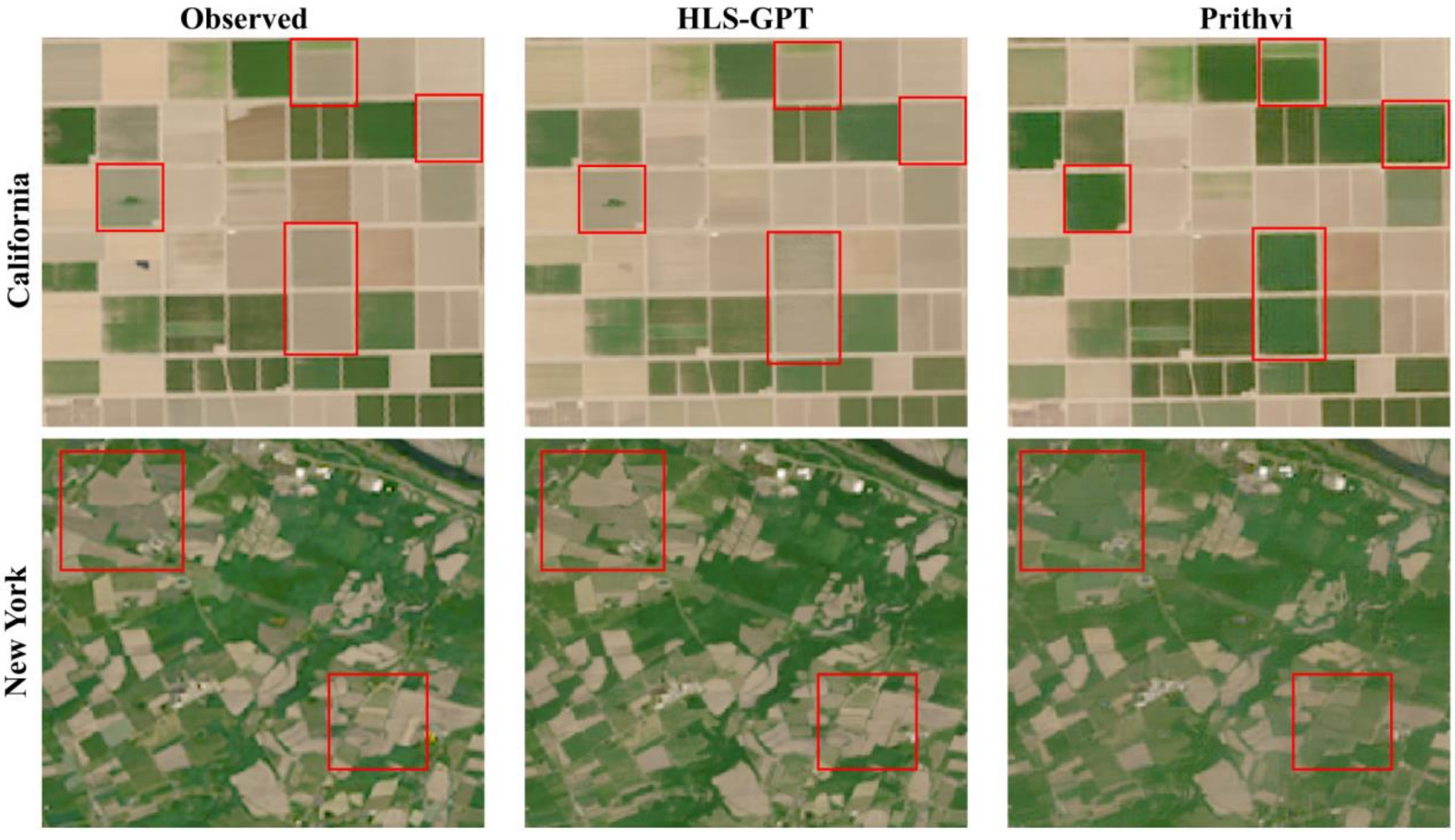

**Fig. 13.** Detailed 190 × 165 30 m pixel spatial subsets selected from the Fig. 12 California and New York tile results. The red boxes highlight regions containing significant reconstruction differences between the HLS-GPT and Prithvi models.

Table 4 summarizes band-specific $RMSE_b^{all_tiles}$ and $SSIM_b^{all_tiles}$ values for the six Landsat and Sentinel-2 bands used by the Prithvi model. For the conventional reconstruction methods, the RMSE results indicate that SG consistently outperforms HANTS in the visible bands (blue, green, and red), likely due to its ability to effectively smooth noisy observations while preserving local variations. In contrast, HANTS shows slightly better performance in the NIR band. In contrast, the two deep learning-based models, particularly HLS-GPT, achieve consistently higher reconstruction accuracy across all six bands, both in terms of spectral fidelity (lower RMSE) and structural consistency (SSIM closer to 1).

**Table 4**. Band-specific $RMSE_b^{all_tiles}$ (Eq. 6) and $SSIM_b^{all_tiles}$ for the HLS-GPT and benchmark method reconstructions derived for each of the six Landsat and Sentinel-2 bands used by Prithvi across the nine evaluation tiles. The bolded values indicate higher tile reconstruction performance (i.e., lower $RMSE_b^{all_tiles}$ or $SSIM_b^{all_tiles}$ closer to 1). The central wavelengths for the Sentinel-2 bands are shown for convenience (note that the 7 Landsat bands have slightly different spectral band passes and central wavelengths but these differences are minimized in the HLS processing).

| Band (central wavelength) | $RMSE_b^{all_tiles}$ | | | | $SSIM_b^{all_tiles}$ | | | |
|---|---|---|---|---|---|---|---|---|
| | HLS-GPT | Prithvi | HANTS | SG filter | HLS-GPT | Prithvi | HANTS | SG filter |
| Blue (0.490 μm) | **0.0052** | 0.0081 | 0.0198 | 0.0138 | **0.9843** | 0.9667 | 0.9141 | 0.9651 |
| Green (0.560 μm) | **0.0059** | 0.0092 | 0.0201 | 0.0145 | **0.9879** | 0.9711 | 0.9358 | 0.9714 |
| Red (0.665 μm) | **0.0076** | 0.0127 | 0.0229 | 0.0168 | **0.9826** | 0.9580 | 0.9190 | 0.9625 |
| NIR (narrow) (0.865 μm) | **0.0120** | 0.0219 | 0.0232 | 0.0251 | **0.9704** | 0.9320 | 0.9708 | 0.9514 |
| SWIR 1 (1.610 μm) | **0.0146** | 0.0205 | 0.0287 | 0.0244 | **0.9680** | 0.9562 | 0.9052 | 0.9513 |
| SWIR 2 (2.190 μm) | **0.0144** | 0.0170 | 0.0243 | 0.0211 | **0.9726** | 0.9623 | 0.9138 | 0.9560 |

Table 5 summarizes reconstruction $RMSE_{all_bands,class}^{all_tiles}$ values for each of the eight Level-1 land cover classes defined by the year 2023 National Land Cover Database (NLCD) 30 m land cover product (Sohl et al., 2025). No SSIM metric values are shown because land cover boundaries and isolated class pixels disrupt the spatial continuity needed for meaningful SSIM computation as noted in Section 3.5. No results for HANTS and SG are included given their lower reconstruction performance particularly over tiles with complex surfaces (see Table 3). The NLCD class-specific HLS-GPT and Prithvi results in Table 5 show that the HLS-GPT provided generally more accurate reconstruction. The HLS-GPT $RMSE_{all_bands,class}^{all_tiles}$ values were smallest for the natural vegetation classes, i.e., for the Herbaceous (0.0053), Shrubland (0.0057), and Forest (0.0067) classes, whereas the Planted/Cultivated cropland class had the highest value (0.0169) likely due to the often complex temporal reflectance changes over croplands (e.g., Fig. 13) associated with growth and harvesting. The HLS-GPT model had lower reconstruction errors than Prithvi for *all* the land cover classes except the water class which had the highest HLS-GPT $RMSE_{all_bands,class}^{all_tiles}$ class value (0.0161) (the Prithvi $RMSE_{all_bands,class}^{all_tiles}$ water class value was 0.0135). This may reflect the fundamental difference between using reconstruction models with spatial (Prithvi) rather than single pixel (HLS-GPT) time series information over spatially more coherent but temporally noisy water surfaces and is discussed further in Section 5.

**Table 5.** Land cover class specific reconstruction $RMSE_{all_bands,class}^{all_tiles}$ (Eq. 5) for the HLS-GPT and Prithvi reconstructions (derived for six Landsat and Sentinel-2 bands used by Prithvi) for each of the 8 NLCD land cover classes. The number (and relative proportion) of pixels for each class are summarized. The bolded values indicate higher tile reconstruction performance (lower $RMSE_{all_bands,class}^{all_tiles}$ ).

| NLCD class | $RMSE_{all_bands,class}^{all_tiles}$ | | Number of pixels in all 9 tiles |
|---|---|---|---|
| | HLS-GPT | Prithvi | |
| Water | 0.0161 | **0.0135** | 1,906,318 (1.9%) |
| Developed | **0.0104** | 0.0172 | 6,850,952 (6.9%) |
| Barren | **0.0151** | 0.0186 | 3,167,530 (3.2%) |
| Forest | **0.0067** | 0.0120 | 29,879,346 (30.0%) |
| Shrubland | **0.0057** | 0.0095 | 16,215,886 (16.3%) |
| Herbaceous | **0.0053** | 0.0100 | 12,871,695 (12.9%) |
| Planted/Cultivated | **0.0169** | 0.0237 | 24,092,164 (24.2%) |
| Wetlands | **0.0101** | 0.0171 | 4,534,125 (4.6%) |

## 5. Discussion

### 5.1 Advantages of HLS-GPT over the NASA-IBM Prithvi model

The key distinction between the HLS-GPT and the NASA–IBM Prithvi models lie in their use of spatial and temporal information for reconstruction. The HLS-GPT is a fully temporal model that reconstructs reflectance using single-pixel 12-month time series, whereas Prithvi primarily exploits spatial information from image patches. Our results demonstrate that leveraging temporal information yielded lower reconstruction errors than relying mainly on spatial context. Notably, the HLS-GPT model did not degrade the spatial reconstruction performance, as assessed by the SSIM metric and by visual examination, even though the reflectance was reconstructed at each pixel location independently of the neighboring pixels. The use of single pixel time series is entirely consistent with previous remote sensing methods, developed prior to deep learning, for example, that fit single pixel time series models, including logistic (Zhang et al., 2003; Jönsson et al., 2018), harmonic (Zhu et al., 2015; Roy and Yan, 2020), and asymmetric Gaussian (Jonsson & Eklundh 2002; Hurley et al., 2014), to take advantage of the seasonal temporal variation of vegetated surfaces.

The NASA–IBM Prithvi was trained using spatial and spectral information defined in 224×224 30 m pixel HLS patches with four-patch sequences acquired within a two year period over the same geographic location (Szwarcman et al., 2025). In certain circumstances spatial information may be more useful for predicting reflectance than the temporal information at single pixel locations. For example, surfaces with spatially structured features, such as built infrastructure, or homogenous surfaces, such as water, are likely to provide spatial contextual information that can be learned more efficiently and reliably than natural surfaces such as forests composed of spatially random stands of trees. This may partially explain why of the eight land cover classes, the Prithvi model provided superior reconstruction error for only the water class (Table 5). As illustrated in Fig. 10, single pixel reflectance time series over water can be temporally "noisy" due to factors including atmospheric correction errors, undetected cloud contamination, and temporal variations in the water constituents, but these factors may be similar across individual 224×224 30 m pixel patches. This suggests that development of a model based on the ViT-H (huge Vision Transformer) architecture or similar, like Prithvi, but using (i) many temporal patch sequences and (ii) some form of random temporal training strategy such as the masked temporal reconstruction (MTR) training strategy we developed, could be a fruitful future research avenue. In fact, at the time of writing this paper, Google's AlphaEarth Foundations model (Brown et al., 2025) was released in August 2025 and employed ViT-based architectures trained on one year of 10 m time series patches (128 × 128 pixels) globally sampled from Landsat 8/9 optical and thermal, Sentinel-2 optical, Sentinel-1 C-band Synthetic Aperture Radar (SAR), Phased Array type L-band SAR-2 (PALSAR-2), digital elevation model (DEM), and meteorological data using a self-supervised pretraining strategy. AlphaEarth adopted a masking approach similar to ours, except that it masked an entire patch at a time for a given sensor source. Another difference is that AlphaEarth supplied cloud masks together with cloud-contaminated observations as input predictors rather than explicitly excluding cloud-contaminated observations as we did for HLS-GPT. To date, Google has released only the global embedding dataset (i.e., final-layer feature vectors) generated by the AlphaEarth model for downstream tasks such as land-cover mapping and biomass estimation, while the foundation model itself remains unavailable, preventing direct comparison with our results.

Another advantage of the HLS-GPT model is that, unlike the NASA–IBM Prithvi model, it can reconstruct the Sentinel-2 red-edge bands that are not acquired by Landsat. The red-edge (~705–740 nm) may detect vegetation stress earlier and more sensitively than red (~660 nm) and NIR (>700 nm) wavelengths (Horler et al., 1983; Gitelson et al., 1994). The Sentinel-2 red-edge bands have proven useful for chlorophyll content estimation (Delegido et al., 2011), land cover mapping (Forkuor et al., 2018), crop nitrogen assessment (Clevers & Gitelson, 2013), gross primary productivity (Lin et al., 2019), and biomass burning discrimination (Huang et al., 2016). Previous efforts reconstructed red-edge bands on Landsat dates using spectral relationships with other Sentinel-2 bands (Scheffler et al., 2020; Shen & Shao, 2024; Mukherjee & Liu, 2023) and previous studies used red-edge BRDF model parameters needed to derive Sentinel-2 red-edge NBAR by a linear wavelength interpolation of MODIS red and NIR BRDF model parameters (Roy et al., 2017). In this study we demonstrated that the HLS-GPT reconstruction error for the red-edge bands was comparable to the NIR bands, suggesting that red-edge reconstruction quality is less affected by the absence of Landsat red-edge bands. This arises because the HLS-GPT model jointly exploits both spectral and temporal relationships through two sensor-specific Transformers modeling the Landsat and Sentinel-2 time series independently, and a cross-sensor Transformer fusing their representations (see Fig. 4a).

Unlike the NASA–IBM Prithvi model, the HLS-GPT can generate 12-month time series starting from any day of the year, potentially enabling, with modification, near real-time medium resolution applications such as within-season crop type mapping (Zhang et al., 2025a) and temporally rolling change detection such as deforestation alerts (Hansen et al., 2016), phenology event detection (Tran et al., 2025), and burned area mapping (Roy et al., 2019). This property enables the HLS-GPT to reconstruct reflectance across years. This may be useful for many applications. For example, for crops such as winter wheat and rapeseed that span calendar years in the northern hemisphere, random starting points allows the model to ingest “crop-year” rather than calendar year sequences, avoiding errors caused by splitting critical growth stages across the boundaries of calendar years. Similarly, certain regions have fire seasons that start and finish before and after December and January, respectively (Boschetti and Roy, 2008). We note also that the HLS-GPT leverages all available good-quality observations to capture the full seasonal dynamics of the land surface. In contrast, the Prithvi model requires selecting mostly clear (<20% cloud cover) tiles, which can be difficult to obtain in persistently cloudy regions.

## 5.2 HLS-GPT's ability to capture complex seasonality and abrupt changes under sparse observation conditions

The HLS-GPT model performed well reconstructing reflectance time series with simple dynamics, e.g., one greenness peak with temporally rich (Fig. 6) and sparse (Fig. 7) time series, and also quite complex dynamics, including multiple greenness peaks (Fig. 8) and rapid surface change (Fig. 9). It is well established that complex surface dynamics often cannot be represented when fitting single pixel time series models (such as logistic and harmonic), particularly when the observations are temporally sparse or irregular (Brooks et al., 2012; Roy and Yan, 2020; Zhang et al., 2021). In contrast, the Transformer architecture used by the HLS-GPT makes no assumptions of reflectance stationarity, periodicity, or symmetry that are implicit in conventional single pixel time series fitting models. To further examine the robustness of the HLS-GPT model to using sparse observation time series, Fig. 14 shows the complex alfalfa single pixel time series (first shown in Fig. 8) with progressively fewer observations that were removed from the eight harvesting (i.e., low NDVI) periods in 2023 (denoted by green boxes). Specifically, Fig. 14(a) shows the observed (black) and reconstructed (orange) NDVI time series as Fig. 8, which contains eight clear V-shaped patterns, each associated with a harvest–regrowth cycle. Figs. 14(b–f) illustrate the reconstructed NDVI after excluding 1, 2, 3, 4, to 5, respectively, of the lowest-NDVI value observations from each trough. Remarkably, even without a post-harvest observation in four of the eight troughs (the third, fourth, fifth and sixth troughs) in Fig. 14(f), the HLS-GPT was still able to reconstruct consistent and physically reasonable V-shape dynamics, effectively capturing the harvest–regrowth transitions solely from the growth-phase signals. Although the reconstructed post-harvest NDVI tends to be higher than the true trough values, the model reliably reproduced the overall temporal pattern. This highlights that HLS-GPT has a capacity for deep, high-dimensional feature synthesis and representation. This capacity is required to infer complex phenological behavior needed to reconstruct plausible reflectance trajectories, and further demonstrates the potential application of HLS-GPT beyond reconstruction tasks as discussed above.

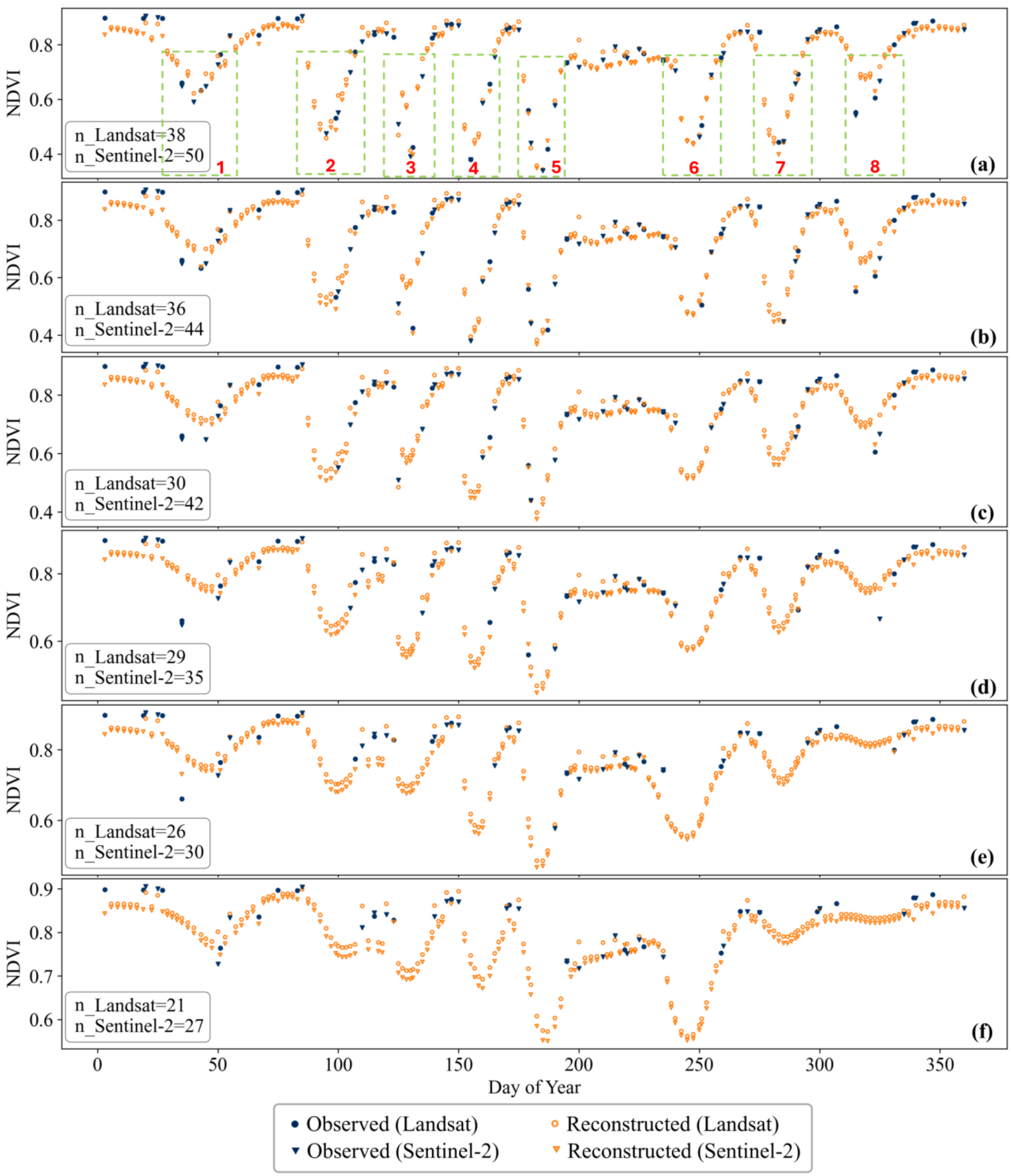

**Fig. 14**. Time series NDVI reconstruction values (orange) for the Fig. 8 California alfalfa single pixel time series, reconstructed after removing progressively more observations from the eight harvesting (i.e., low NDVI) periods in 2023 (denoted by green boxes). The n_Landsat and n_Sentinel-2 values denote the number of good-quality HLS processed Landsat and Sentinel-2 observations (black) in 2023. The results in (a) are the same as the NDVI values shown in Fig. 8. The NDVI is derived from the reconstructed red and NIR reflectance values.

### 5.3 Future suggested work

The HLS-GPT model inference requires ~66 minutes on a GPU node with a single NVIDIA A100 80GB card to reconstruct daily reflectance for all the HLS bands for a 12 month period. The total end-to-end runtime is typically 170 minutes and is dominated by the HLS tile data input/output. Further computational efficiency gains could be achieved by optimizing the input/output and perhaps also by using a more computationally efficient deep learning architecture, such as Sparse Transformer (Child et al., 2019) or Mamba (Gu and Dao, 2023).

The HLS-GPT model can be scaled globally without any structure modification. Despite this, one challenge for scaling lies in the diversity of phenological regimes across global biomes. For instance, tropical ecosystems exhibit weak or unseasonal phenological cycles, while mid- and high-latitude regions show strong seasonality. In the Southern Hemisphere, phenology is phase-shifted relative to the Northern Hemisphere, further complicating the representation of global vegetation dynamics. These regional contrasts raise an open question: can a single unified model effectively capture global phenological diversity, or would multiple region-specific models be required to better represent distinct climatic and ecological domains?

The HLS-GPT can be also considered as a pre-trained foundation model as noted in the introduction and could be fine-tuned for downstream time series tasks such as crop type mapping, forest monitoring, disaster response, burned area mapping, and flood detection. Recent years have witnessed a surge of interest in Geospatial Foundation Models (GFMs) (Cong et al., 2022; Reed et al., 2023; Fuller et al., 2023; Xiong et al., 2024; Marsocci et al., 2024; Qin et al., 2025; Liu et al., 2025; Yu et al., 2025). By performing self-supervised pretraining on millions of HLS samples across CONUS, the HLS-GPT learns rich feature representations that can be readily transferred to downstream applications. Section 5.2 has already demonstrated the strong latent feature integration and representational capacity of the HLS-GPT. A notable distinction is that most existing temporally aware GFMs—such as Prithvi-EO-2.0 (Szwarcman et al., 2025), TiMo (Qin et al., 2025), RobSense (Do et al., 2025), and AgriFM (Li et al., 2025b)—are constrained to a limited number of timestamps, which restricts their ability to capture phenological diversity. In contrast, HLS-GPT encompasses an entire annual cycle with arbitrary starting points and incorporates every good-quality observation to maximize use of surface dynamics information. In addition to modelling dynamics, single-pixel time series modelling also aligns more closely with real-world conditions where temporal sampling are irregular or unaligned across sensors and acquisition dates as we have discussed earlier and has been noted by Dumeur et al. (2025) and Zhang et al. (2025a). In future work, we will extend this paradigm to systematically evaluate how HLS-GPT pretrained weights can improve downstream tasks under limited sample availability, thereby bridging the gap between time series large-scale pretraining and practical Earth observation applications, including, for example, crop type mapping, burned area mapping, soil moisture mapping, and flood monitoring, particularly under conditions with limited labeled data.

## 6. Conclusion

In this study, we developed and presented the HLS-GPT model for 12-month single pixel time series reflectance reconstruction. The extensive training set (104 million Landsat and Sentinel-2 good-quality observations systemically sampled across the CONUS from 9 years of HLS data) combined with the MTR training strategy enabled the model to learn diverse patterns from the data themselves. The HLS-GPT model provides reflectance reconstruction for all 7 Landsat bands and all 11 Sentinel-2 bands including the red-edge bands. The MTR training strategy is self-supervised and generates 12 month time series with arbitrary starting dates to ensure learning of complex temporal dynamics to help capture reflectance seasonality, phenological patterns, and inter-annual variations.

The HLS-GPT model achieved high reconstruction accuracy when evaluated using test pixel reflectance time series which were not used to train the model. For Landsat, the RMSE ranged from 0.0147 in the coastal blue band to 0.0240 in the SWIR 1 band, and the rRMSE ranged from 8.98% in the NIR band to 30.52% in the coastal blue band. For Sentinel-2, the RMSE ranged from 0.0142 in the blue band to 0.0259 in SWIR 1 band, and the rRMSE ranged from 9.29% in the narrow NIR band to 34.03% in the coastal blue band. These rRMSE values are relatively low given that the relative mean difference between same-day Landsat–Sentinel-2 HLS processed reflectance was 4% for the red, NIR and SWIR bands and up to 10.5% for the coastal blue band (Ju et al., 2025). Reconstruction experiments on nine independent evaluation 109×109 km HLS tiles demonstrated the efficacy of the HLS-GPT model reconstruction and its superior performance over both conventional time series fitting methods and the recent NASA–IBM Prithvi version 2.0 model (Szwarcman et al., 2025), achieving better RMSE and SSIM metrics for each tile and each band. The improvement was more pronounced over agricultural land, where surface dynamics are highly complex due to frequent field management activities (e.g., tillage, seeding, irrigation, fertilizer application and harvesting). Case studies of alfalfa harvest-regrowth cycles and wildfire disturbance demonstrate that the HLS-GPT can robustly model both complicated seasonality and abrupt changes, while it can also reconstruct unobserved temporal shapes by inferring pixel-specific temporal reflectance variations.

In summary, the HLS-GPT ingests good-quality-flagged HLS time series that, due to a number of factors, have irregular observation temporal cadence and outputs daily Landsat and Sentinel-2 reflectance. To facilitate user application, the application codes are publicly available at https://github.com/hankui/HLS-GPT and the trained model and training samples are available at https://zenodo.org/records/15678251. The open-source pretrained HLS-GPT model can be readily used for time series reflectance reconstruction that can then be used to support diverse medium resolution time series applications.

**Acknowledgements**

This work was supported by the NASA Advanced Information Systems Technology (AIST) (grant 80NSSC25K7779), NASA Advancing Collaborative Connections for Earth System Science (ACCESS) (grant 80NSSC21M0023), and NASA Established Program to Stimulate Competitive Research (EPSCoR) (grant 80NSSC24M0116) programs.

## Appendix A. Summary of Deep Learning-Based Reconstruction Studies Using Single Data Sources

**Table A1**. A summary of deep learning-based reconstruction studies using single data sources or harmonized products (e.g., HLS). The Input2Output column indicates whether the model is patch-based or single-pixel–based, and for patch-based models, it also reports the patch size. Note the studies using additional auxiliary data, including SAR data (e.g., Meraner et al., 2020; Darbaghshahi et al., 2021; Peng et al., 2022; Xu et al., 2022a; Zhao et al., 2023; Dai et al., 2025; Shu et al., 2025; Wang et al., 2026) and MODIS data (e.g., Liu et al., 2019; Yang et al., 2021; Ma et al., 2023; Gu et al., 2023; Guo et al., 2024; Zhao et al., 2025; Guo et al., 2025; Wu et al., 2025; Irigireddy and Bandaru, 2025) were not included in this table.

| Literature | Study area (Training data) | Input2Output | Sequence length of training data | Reconstruction Bands | Model published |
|---|---|---|---|---|---|
| Zhang et al., 2020 | 7 images in Australia | Patch (40 × 40) Seq2I | 4 time steps over 4 months | Landsat/Sentinel-2 (RGB+NIR) | √ |
| ST-net (Chen et al., 2020) | 20 images in Asia and Oceania | Patch (256 × 256) Seq2I | 16 days interval | Landsat (Coastal, RGB, NIR, SWIR 1/2) | × |
| AMGAN-CR (Xu et al., 2022b) | 9 images in Australia | Patch (256 × 256) I2I | Single | Landsat (Coastal, RGB, NIR, SWIR 1/2) | × |
| U-TILISE (Stucker et al., 2023) | EarthNet2021 (Central and Western Europe) (Requena-Mesa et al., 2021) | Patch (128 × 128) Seq2Seq | 10 time steps (5 days interval) | Sentinel-2 (RGB+NIR) | √ |
| SSLI (Liu et al., 2024) | 5 regions in CONUS | Single pixel Seq2Seq | 122 time steps over 1 calendar year | HLS (Coastal, RGB, NIR, SWIR 1/2) | × |
| RFE-VCR (Jin et al., 2024) | 10 Jilin-1 satellite videos in 10 global locations | Patch (320 × 320) Seq2Seq | 8 frames | RGB | × |
| GANFilling (Gonzalez-Calabuig et al., 2025) | EarthNet2021 (Central and Western Europe) (Requena-Mesa et al., 2021) | Patch (128 × 128) Seq2Seq | 10 time steps (5 days interval) | Sentinel-2 (RGB+NIR) | √ |
| Prithvi-EO-2.0 (Szwarcman, et al., 2025) | Global | Patch (256 × 256) Seq2Seq | 4 time steps with intervals of 1 to 6 months | HLS (RGB, NIR, SWIR 1/2) | √ |
| TAMGAN (Wang et al., 2025b) | 8 MODIS tiles in US | Patch (64 × 64) Seq2Seq | 46 time steps over 1 calendar year | MODIS (RGB+NIR) | × |
| MST-Net (Wang et al., 2025a) | 13 regions in global | Patch (40 × 40) Seq2I | 5 time steps | Landsat (RGB, NIR, SWIR 1/2) | × |
| SSA-BiTCN-Attention (Zhang et al., 2025b) | Global | Single pixel Seq2I | 114 time steps over 5 years | Landsat FAPAR | × |
| CloudRuler (Li et al., 2025a) | 20 images in Southeast CONUS | Patch (256 × 256) I2I | Single | Landsat (Optical + TIR) | √ |

**Appendix B. Dates of input images for the Prithvi model.**

**Table B1**. Selected input and reconstruction dates (yyyyddd, where yyyy represents the four-digit year and ddd the three-digit day of year (DOY)) for the Prithvi model across nine HLS test tiles.

| Tile ID | Selected dates | Reconstruction date |
|---|---|---|
| 11SPS | 2023187, 2023190, 2023200 | 2023195 |
| 14TNP | 2023122, 2023127, 2023142 | 2023140 |
| 12TVL | 2023268, 2023272, 2023292 | 2023282 |
| 11TQM | 2023213, 2023223, 2023238 | 2023226 |
| 13SEC | 2023263, 2023265, 2023273 | 2023268 |
| 15SWB | 2023063, 2023073, 2023088 | 2023078 |
| 17RNL | 2023011, 2023015, 2023045 | 2023019 |
| 18TWN | 2023132, 2023135, 2023150 | 2023148 |
| 10TDM | 2023177, 2023180, 2023187 | 2023182 |